\pdfoutput=1

\documentclass[11pt]{article}

\usepackage[preprint]{acl}

\usepackage{times}
\usepackage{latexsym}
\usepackage{graphicx}
\usepackage{xspace, soul}
\usepackage{threeparttable}
\usepackage{booktabs}
\usepackage{multirow}
\usepackage{amsmath}
\usepackage{cleveref}
\usepackage{pifont}
\usepackage{amssymb,enumitem}
\usepackage{comment}
\usepackage{hyperref,graphicx}
\usepackage{marvosym}
\usepackage{booktabs, makecell, rotating, siunitx}
\usepackage{tcolorbox}  
\usepackage{listings}   
\usepackage{xcolor}     
\usepackage{colortbl}
\usepackage{subcaption}
\usepackage{longtable} 
\usepackage{pdflscape} 
\usepackage{tcolorbox}
\usepackage[ruled,vlined]{algorithm2e}
\usepackage[T1]{fontenc}

\usepackage[utf8]{inputenc}

\usepackage{microtype}

\usepackage{inconsolata}

\usepackage{graphicx}

\title{Large Language Models for Controllable Multi-property Multi-objective Molecule Optimization}

\author{
 \textbf{Vishal Dey\textsuperscript{1}},
 \textbf{Xiao Hu\textsuperscript{1}},
 \textbf{Xia Ning\textsuperscript{1,2,3,4}}
\\
 \textsuperscript{1} Department of Computer Science
and Engineering, The Ohio State University, USA \\
 \textsuperscript{2} Translational Data Analytics Institute, The Ohio State University, USA \\
 \textsuperscript{3}Department of Biomedical Informatics, The Ohio State University, USA \\
 \textsuperscript{4} College of Pharmacy, The Ohio State University, USA
\\
 \small{
   \textbf{Correspondence:} \href{mailto:email@domain}{ning.104@osu.edu}
 }
}

\newcommand{\cmark}{\ding{51}}%
\newcommand{\xmark}{\ding{55}}%
\definecolor{lightgray}{gray}{0.9} 
\definecolor{lightergray}{gray}{0.95} 

\newcommand{\PropImpv}{\mbox{$\mathop{\mathtt{\mathcal{P}_{i}}}\limits$}\xspace}
\newcommand{\PropStable}{\mbox{$\mathop{\mathtt{\mathcal{P}_{s}}}\limits$}\xspace}
\newcommand{\DrugMOpt}{\mbox{$\mathop{\mathtt{MolOpt\text{-}Instructions}}\limits$}\xspace}
\newcommand{\OptData}{\mbox{$\mathop{\mathtt{MuMOInstruct}}\limits$}\xspace}
\newcommand{\MOptData}{\mbox{$\mathop{\mathtt{C\text{-}MuMOInstruct}}\limits$}\xspace}
\newcommand{\GTO}{\mbox{$\mathop{\mathtt{GT}}\limits$}\xspace}
\newcommand{\CSO}{\mbox{$\mathop{\mathtt{CS}}\limits$}\xspace}

\newcommand{\mollm}{\mbox{$\mathop{\mathtt{GeLLM^4O\text{-}C}}\limits$}\xspace}
\newcommand{\mollmNTask}{\mbox{$\mathop{\mathtt{GeLLM^4O\text{-}C\text{-}N}}\limits$}\xspace}
\newcommand{\mollmNGen}{\mbox{$\mathop{\mathtt{GeLLM^4O\text{-}C\text{-}P(N)}}\limits$}\xspace}
\newcommand{\mollmDecGen}{\mbox{$\mathop{\mathtt{GeLLM^4O\text{-}C\text{-}P(10)}}\limits$}\xspace}

\newcommand{\mollmTaskM}{\mbox{$\mathop{\mathtt{GeLLM^4O\text{-}C\text{-}N_{Mistral}}}\limits$}\xspace}
\newcommand{\mollmTaskL}{\mbox{$\mathop{\mathtt{GeLLM^4O\text{-}C\text{-}N_{Llama}}}\limits$}\xspace}

\newcommand{\mollmTripleTaskM}{\mbox{$\mathop{\mathtt{GeLLM^4O\text{-}C\text{-}{3}_{Mistral}}}\limits$}\xspace}
\newcommand{\mollmQuadTaskM}{\mbox{$\mathop{\mathtt{GeLLM^4O\text{-}C\text{-}{4}_{Mistral}}}\limits$}\xspace}
\newcommand{\mollmTripleTaskL}{\mbox{$\mathop{\mathtt{GeLLM^4O\text{-}C\text{-}{3}_{Llama}}}\limits$}\xspace}
\newcommand{\mollmQuadTaskL}{\mbox{$\mathop{\mathtt{GeLLM^4O\text{-}C\text{-}{4}_{Llama}}}\limits$}\xspace}

\newcommand{\mollmNGenM}{\mbox{$\mathop{\mathtt{GeLLM^4O\text{-}C\text{-}P(N)_{Mistral}}}\limits$}\xspace}
\newcommand{\mollmTripleGenM}{\mbox{$\mathop{\mathtt{GeLLM^4O\text{-}C\text{-}P(3)_{Mistral}}}\limits$}\xspace}
\newcommand{\mollmQuadGenM}{\mbox{$\mathop{\mathtt{GeLLM^4O\text{-}C\text{-}P(4)_{Mistral}}}\limits$}\xspace}
\newcommand{\mollmDecGenM}{\mbox{$\mathop{\mathtt{GeLLM^4O\text{-}C\text{-}P(10)_{Mistral}}}\limits$}\xspace}

\newcommand{\mollmNGenL}{\mbox{$\mathop{\mathtt{GeLLM^4O\text{-}C\text{-}P(N)_{Llama}}}\limits$}\xspace}
\newcommand{\mollmTripleGenL}{\mbox{$\mathop{\mathtt{GeLLM^4O\text{-}C\text{-}P(3)_{Llama}}}\limits$}\xspace}
\newcommand{\mollmQuadGenL}{\mbox{$\mathop{\mathtt{GeLLM^4O\text{-}C\text{-}P(4)_{Llama}}}\limits$}\xspace}
\newcommand{\mollmDecGenL}{\mbox{$\mathop{\mathtt{GeLLM^4O\text{-}C\text{-}P(10)_{Llama}}}\limits$}\xspace}

\newcommand{\BPQ}{\mbox{$\mathop{\mathtt{BPQ}}\limits$}\xspace}
\newcommand{\ELQ}{\mbox{$\mathop{\mathtt{ELQ}}\limits$}\xspace}
\newcommand{\ACEP}{\mbox{$\mathop{\mathtt{ACEP}}\limits$}\xspace}
\newcommand{\BDPQ}{\mbox{$\mathop{\mathtt{BDPQ}}\limits$}\xspace}
\newcommand{\DHMQ}{\mbox{$\mathop{\mathtt{DHMQ}}\limits$}\xspace}
\newcommand{\CDE}{\mbox{$\mathop{\mathtt{CDE}}\limits$}\xspace}
\newcommand{\ABMP}{\mbox{$\mathop{\mathtt{ABMP}}\limits$}\xspace}
\newcommand{\BCMQ}{\mbox{$\mathop{\mathtt{BCMQ}}\limits$}\xspace}
\newcommand{\BDEQ}{\mbox{$\mathop{\mathtt{BDEQ}}\limits$}\xspace}
\newcommand{\HLMPQ}{\mbox{$\mathop{\mathtt{HLMPQ}}\limits$}\xspace}

\newcommand{\LlaSMol}{\mbox{$\mathop{\mathtt{LlaSMol}}\limits$}\xspace}
\newcommand{\LlaSMolM}{\mbox{$\mathop{\mathtt{LlaSMol_{Mistral}}}\limits$}\xspace}
\newcommand{\ChemDFM}{\mbox{$\mathop{\mathtt{ChemDFM}}\limits$}\xspace}
\newcommand{\ChemDFML}{\mbox{$\mathop{\mathtt{ChemDFM_{Llama}}}\limits$}\xspace}

\newcommand{\SR}{\mbox{$\mathop{\mathtt{SR}}\nolimits$}}
\newcommand{\SSR}{\mbox{$\mathop{\mathtt{SR_\Theta}}\nolimits$}}
\newcommand{\Val}{\mbox{$\mathop{\mathtt{Val}}\nolimits$}}
\newcommand{\SAS}{\mbox{$\mathop{\mathtt{SAS}}\nolimits$}}
\newcommand{\Sim}{\mbox{$\mathop{\mathtt{Sim}}\nolimits$}}
\newcommand{\Nov}{\mbox{$\mathop{\mathtt{Nov}}\nolimits$}}

\newcommand{\RI}{\mbox{$\mathop{\mathtt{RI}}\nolimits$}}
\newcommand{\APS}{\mbox{$\mathop{\mathtt{APS}}\nolimits$}}

\newcommand{\ImpT}{\mbox{$\mathop{\mathtt{Impv\text{-}Spec}}\limits$}\xspace}
\newcommand{\ImpG}{\mbox{$\mathop{\mathtt{Impv\text{-}Gen}}\limits$}\xspace}

\begin{document}
\maketitle

\begin{abstract}
%
%
In real-world drug design, 
molecule optimization requires selectively improving
multiple molecular properties up to pharmaceutically relevant levels,
while maintaining others that already meet such criteria. 
However, existing computational approaches and instruction-tuned LLMs 
fail to capture such nuanced property-specific objectives, 
limiting their practical applicability.
To address this, we introduce \MOptData, the first
instruction-tuning dataset focused on multi-property
optimization with explicit, property-specific objectives.
Leveraging \MOptData, we develop {\mollm}s, a series of
instruction-tuned LLMs that can perform targeted property-specific optimization.
Our experiments across 5 in-distribution and 5 out-of-distribution tasks
show that {\mollm}s consistently outperform strong baselines,
achieving up to 126\% higher success rate.
Notably, {\mollm}s exhibit impressive 0-shot generalization
to novel optimization tasks and unseen instructions.
This offers a step toward a foundational LLM
to support realistic, diverse optimizations with property-specific objectives.
\MOptData and code are accessible through 
\url{https://github.com/ninglab/GeLLMO-C}.

\end{abstract}

\section{Introduction}

Developing a new drug is a time-consuming and
expensive process, 
requiring over a decade and \$2 billions~\cite{sertkaya2024costs}.
A key stage in this process is lead optimization~\cite{nicolaou2013},
where ``hit" molecules -- exhibiting promising early-stage bioactivity against drug targets --
are optimized for multiple molecular properties~\cite{Nicolotti2011}
critical for pharmaceutical success.
In practice, this stage often requires 
improving specific properties up to a pharmaceutically significant level,
while maintaining already desirable ones within acceptable bounds.
We refer to this setting 
as controllable multi-property, multi-objective optimization (C-MuMO), 
allowing for property-specific objectives, 
and thus greater control over the optimization.

Such controllable optimization 
requires navigating complex trade-offs among multiple properties
that are often competing or even conflicting~\cite{niu2024tradingoff}.
For instance, optimizing an oral antipsychotic drug
requires sufficiently high blood-brain barrier permeability (BBBP)~\cite{Pollak2018}
and dopamine receptor D2 (DRD2) inhibition~\cite{Seeman2001} 
to access the central nervous system (CNS) and block dopamine receptors in the CNS~\cite{Seeman1976}.
Meanwhile, properties related to toxicity, such as
Potassium (K$^+$) channel inhibition must be lowered,
since excessive inhibition of K$^+$ channels in the brain~\cite{Shepard2007} can cause fatal cardiac arrythmias~\cite{Sanguinetti2006}.
%
Additionally, properties supporting oral bioavailability, such as intestinal absorption, 
must be maintained if they already meet desirable levels. 
These trade-offs highlight the need for property-specific objectives to mimic realistic optimization tasks.

Most existing computational approaches~\cite{gao2022sample,jensen2019graph,you2018graph,blaschke2020reinvent} 
cannot handle tasks with multiple objectives. 
%
Furthermore, existing approaches for multi-objective optimization~\cite{sun2022molsearch,kim2024genetic,wu2024leveraging} rely on manually curated reward functions and careful task-specific tuning --
limiting their scalability and applicability to diverse tasks in practice. 
We refer readers to Appendix~\ref{sec:related} for a detailed review of existing approaches.
%
%
Recently, instruction-tuned LLMs~\cite{dey2025gellmo},
demonstrated strong performance on diverse multi-property optimization tasks.
However, they only tackle tasks where all properties should be improved simultaneously.
This setting fails to capture the nuanced property-specific objectives prevalent in realistic lead optimization. 


\begin{figure*}[h!]
    \centering
    \includegraphics[width=\textwidth]{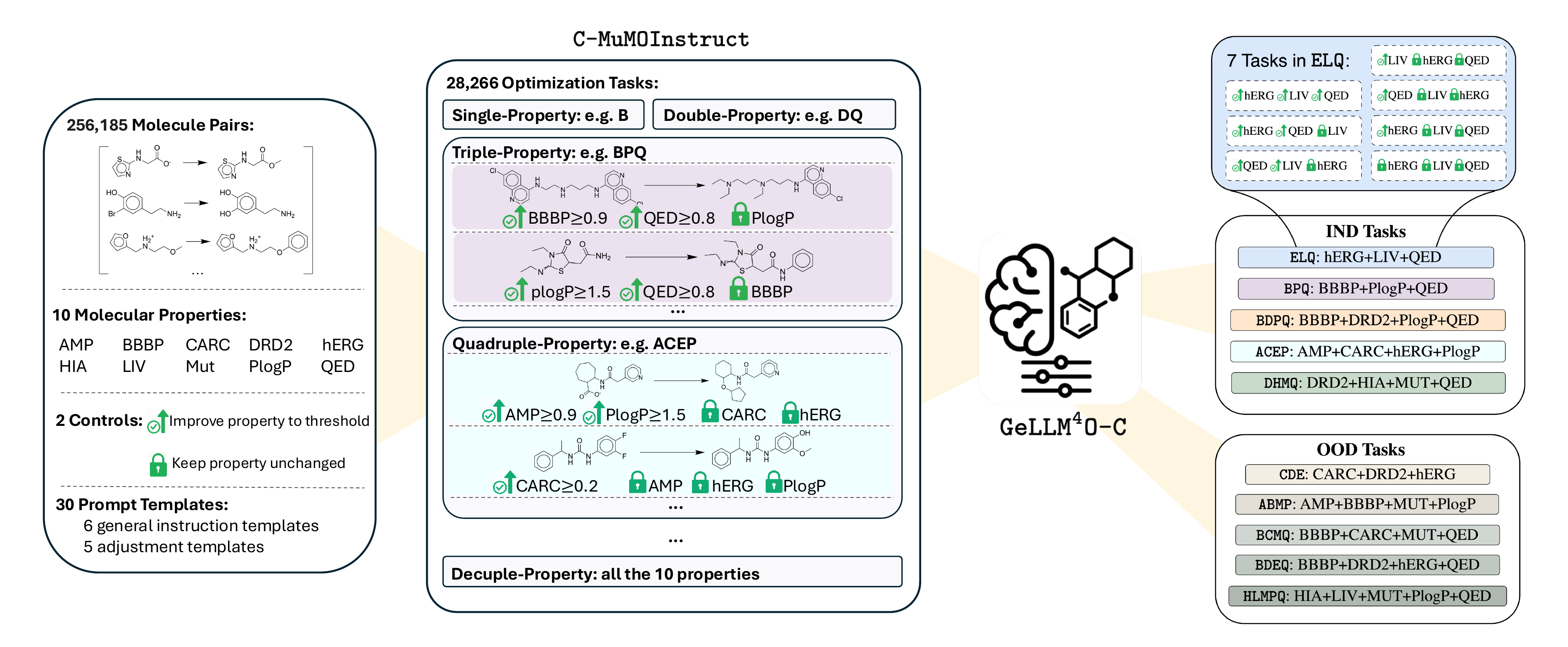}
    \vspace{-25pt}
    \caption{Overview of \MOptData and \mollm}
    \label{fig:Overview}
\vspace{-15pt}
\end{figure*}

To address these critical limitations, 
we introduce \MOptData, 
the first high-quality instruction-tuning dataset designed for C-MuMO tasks
involving up to 10 molecular properties.
%
Unlike prior datasets that require all properties to improve, 
\MOptData explicitly incorporates controllable property-specific objectives
-- 
specifying which properties must be improved up to a user-defined
property-specific threshold,
and which must be maintained within acceptable bounds. 
This design better reflects real-world lead optimization, 
where some properties reach pharmaceutically significant 
levels in early stages, while others require multiple iterations for further improvement.
%

Built on \MOptData, we introduce a family of 
\underline{Ge}neralizable \underline{L}arge \underline{L}anguage \underline{M}odels for \underline{M}ulti-property, \underline{M}ulti-\underline{O}bjective \underline{C}ontrollable optimization,
\mollm,
by instruction-tuning general-purpose LLMs.
%
\mollm is trained to handle tasks requiring selective 
improvement of specific properties while maintaining 
already desirable ones.
We develop both specialist and generalist variants.
Each specialist {\mollm} is trained on a single property combination
with multiple controllable multi-objective tasks.
Generalist {\mollm} is trained across
diverse multi-property combinations and multiple controllable
objectives within each combination,
enabling cross-task knowledge transfer.
This enables a single foundational model to handle novel and diverse C-MuMO tasks
without task-specific fine-tuning.

We evaluate our \mollm models with strong general-purpose LLMs
and foundational LLMs for chemistry across 5 in-distribution (IND) 
and 5 out-of-distribution (OOD) tasks.
Our results reveal several key findings:
\textbf{(1)} All {\mollm}s substantially 
outperform state-of-the-art baselines on all IND and OOD tasks, 
with gains of up to 126\% over the best baselines. 
\textbf{(2)} Generalist {\mollm}s outperform specialist ones on 4 out of 5 IND tasks, with impressive gains of up to 26\% on challenging tasks.
\textbf{(3)} Generalist {\mollm}s demonstrate remarkable 0-shot generalization to OOD tasks, 
outperforming strong baselines by 27\% on average.

%
To the best of our knowledge,
\MOptData is the first large scale, high-quality instruction-tuning
dataset specifically focused on controllable, multi-objective optimization with up to 10 properties.
Generalist {\mollm}s tuned on \MOptData
demonstrate strong generalization abilities,
which
highlights their strong potential to tackle
unseen, diverse C-MuMO tasks
prevalent in realistic drug design scenarios.
%
%
Figure~\ref{fig:Overview} presents the overall framework of \mollm.
Dataset, models, and code are accessible through
\url{https://github.com/ninglab/GeLLMO-C}.

\begin{table}[t!]
\centering
\caption{Comparison among instruction-tuning datasets}
\label{tbl:relwork_data}
\vspace{-5pt}
\begin{threeparttable}
\begin{scriptsize}
\begin{tabular}{
    @{\hspace{1pt}}p{1.8cm}@{\hspace{0pt}}
    @{\hspace{0pt}}c@{\hspace{1pt}}
    @{\hspace{0pt}}c@{\hspace{1pt}}
    @{\hspace{0pt}}c@{\hspace{1pt}}
}
\toprule
\multirow{2}{*}{Comparison} & \scriptsize{\DrugMOpt}
& \scriptsize{\OptData}
& \scriptsize{\MOptData} \\
& \cite{ye2025drugassist} & \cite{dey2025gellmo} & (ours) \\
\midrule

Multi-objective & \xmark & \xmark & \cmark \\
Threshold-based& \cmark & \xmark & \cmark \\
Realistic & \xmark & \cmark & \cmark
\\
\#properties & 5 & 6 & 10 \\
\#molecules & 1,595,839 & 331,586 & 433,166 \\
\#pairs & 1,029,949 & 255,174 & 256,185 \\
\#Total tasks & 8 & 63 & 28,266 \\
~~\#Tasks $\ge3$ prop & 0 & 42 & 27,401 \\
~~\#Eval $\ge3$ prop & 0 & 10 & 119 \\
~~~~\#IND & 8 & 5 & 51\\
~~~~\#OOD & 0 & 5 & 68\\

\bottomrule
\end{tabular}

\end{scriptsize}
\end{threeparttable}
\vspace{-10pt}
\end{table}

\section{\MOptData}
\label{sec:data}

In this paper,
we introduce \MOptData, 
which provides control over each property objective in
multi-property optimization tasks, unlike existing datasets such as \OptData.
This enables models tuned on \MOptData
to improve specific properties up to a user-defined level,
while maintaining others at already desirable levels
-- a crucial capability that distinguishes \MOptData from existing datasets.
These key differences are highlighted in Table~\ref{tbl:relwork_data}.

\paragraph{Problem Definition:}
A 
C-MuMO task is to modify a
hit molecule $M_x$ 
into an improved lead molecule $M_y$,
via structural modifications on $M_x$,
guided by property-specific objectives
-- controlling which properties to be improved and the extent of such improvement.
Given $\mathcal{P}$ molecular properties,
we define a pharmaceutically relevant level, $\Theta_p$,
for each property $p \in \mathcal{P}$,
%
Accordingly, $p$ is considered near-optimal
if its score in $M_x$ -- denoted as $p(M_x)$ -- is
more desirable than $\Theta_p$ (represented as $p(M_x) \prec \Theta_p$), 
and sub-optimal, otherwise (represented as $p(M_x) \succeq \Theta_p$).
The desirability of each property is determined by the intended pharmaceutical goal,
where either higher or lower property scores 
increase the molecule's likelihood to be a successful drug candidate.
For example, a higher BBBP is desired
for drugs targeting the CNS to ensure their access to the brain,
whereas a lower BBBP is desired
for peripheral targets
to prevent damage to the CNS.
%
%
%


%
Formally, a C-MuMO task optimizing $M_x$ to $M_y$ 
aims to improve all sub-optimal properties $\PropImpv = \{p \in \mathcal{P} | p(M_x) \prec \Theta_p\}$ 
while maintaining all near-optimal properties
$\PropStable = \{p \in \mathcal{P} \mid p(M_x) \succeq \Theta_p\}$
such that:
%
\textbf{(1)} 
$M_y$ remains structurally similar to $M_x$
(similarity constraint);
\textbf{(2)} 
$M_y$ improves upon $M_x$ 
in each sub-optimal property $p \in \PropImpv$ 
by at least a property-specific threshold, $\Delta_p$,
represented as $(M_x \prec_{\Delta_p} M_y)_{\forall p \scriptsize{\in \PropImpv}}$
(property improvement constraint);
and 
\textbf{(3)}
the absolute change from $M_x$ to $M_y$ in each near-optimal property 
$p \in \PropStable$ remains within $\Delta_p$
to ensure such properties with already desirable scores are maintained,
represented as
$(M_x \cong_{\Delta_p} M_y)_{\forall p \scriptsize{\in \PropStable}}$
(property stability constraint).
%

%


\subsection{Design Principles}
\label{sec:data:design}

Following the above definition,
we construct \MOptData,
the first high-quality instruction tuning dataset for
C-MuMO tasks with property-specific objectives.
%
%
Our design of \MOptData is based on 5 key principles:
\vspace{-5pt}
\paragraph{(1) Real-world relevance:}
C-MuMO tasks are widely prevalent in real-world lead optimization, where some properties may already meet desirable levels while others require further improvement.
Each optimization task in \MOptData is carefully curated to reflect nuanced multi-property objectives encountered in real-world drug design.
By combining ADMET properties (e.g., intestinal absorption,
mutagenicity) 
with properties related to specific therapeutic endpoints
(e.g., dopamine receptor and potassium channel inhibition),
\MOptData captures complex and realistic multi-property trade-offs.

\vspace{-5pt}
\paragraph{(2) Controllable multi-property threshold-based optimization:}
Unlike prior datasets such as \OptData,
which enforces the same objective for all properties 
(i.e., `improve all' simultaneously),
\MOptData introduces property-specific objectives
-- specifying sub-optimal properties to improve and
near-optimal ones to maintain -- in addition to `improve all' objectives.
Such property-specific objectives
enables modeling diverse multi-property trade-offs,
thereby capturing more realistic optimization scenarios.
%
Furthermore, \MOptData introduces 
property-specific thresholds,
requiring each sub-optimal property to be improved up to a level considered sufficient
for pharmaceutical success.
This enables models tuned on \MOptData to learn more 
targeted optimization strategies and navigate
nuanced multi-property trade-offs more effectively than models
tuned on datasets lacking finer control.
Meanwhile, learning such nuanced and controllable optimization 
introduces additional modeling challenges,
making \MOptData a more practical and difficult dataset than existing ones.
%

%
%
\vspace{-5pt}
\paragraph{(3) Comprehensive coverage:}
Spanning across 10 pharmacologically relevant molecular properties,
\MOptData covers a wide range of multi-property combinations,
and multi-objective tasks with property-specific objectives for each property combination.
This leads to a comprehensive set of optimization tasks,
better capturing the complexity of real-world drug design.

\vspace{-5pt}
\paragraph{(4) Pairwise optimization:}
Following \OptData,
\MOptData is constructed from molecule pairs 
that satisfy similarity, property improvement, and stability constraints.
This enables models to effectively associate targeted 
structural modifications with property changes.
%
%

%
%
%
%
\vspace{-5pt}
\paragraph{(5) Diverse instructions:}
\MOptData provides diverse natural language instructions for each task with varied phrasings.
This prevents instruction-tuned LLMs from overfitting 
to a specific phrasing,
and enables them to generalize to unseen instructions
-- a crucial capability in practice, where task descriptions can widely vary.

\subsection{Overview of \MOptData Tasks}
\label{sec:data:tasks}

\begin{table*}[h!]
\centering
\caption{Summary of \MOptData Tasks for Evaluation 
}
\label{tbl:task_summary}
\vspace{-5pt}
\begin{small}
\begin{threeparttable}
\begin{tabular}{
    @{\hspace{2pt}}l@{\hspace{2pt}}
    @{\hspace{2pt}}l@{\hspace{2pt}}
    @{\hspace{2pt}}c@{\hspace{2pt}}
    @{\hspace{2pt}}c@{\hspace{2pt}}
    @{\hspace{2pt}}c@{\hspace{2pt}}
    @{\hspace{2pt}}c@{\hspace{2pt}}
    @{\hspace{2pt}}c@{\hspace{2pt}}
    @{\hspace{2pt}}c@{\hspace{2pt}}
    @{\hspace{2pt}}c@{\hspace{2pt}}
    @{\hspace{2pt}}c@{\hspace{2pt}}
    @{\hspace{2pt}}c@{\hspace{2pt}}
    @{\hspace{2pt}}c@{\hspace{2pt}}
    @{\hspace{1pt}}r@{\hspace{2pt}}
    @{\hspace{1pt}}r@{\hspace{2pt}}
    @{\hspace{1pt}}r@{\hspace{2pt}}
    @{\hspace{1pt}}r@{\hspace{2pt}}
    @{\hspace{1pt}}c@{\hspace{2pt}}
}
\toprule
\multirow{4}{*}{Type}
& \multirow{4}{*}{$\mathcal{P}$-Comb} 
& \multicolumn{10}{c}{Properties}
& {\#Pairs}
& {\#Mols}
& {\#Test}
& {\#Tasks}
& {Cat}
\\
\cmidrule(){3-12}
& & AMP{$^\uparrow$}
& BBBP{$^\uparrow$} & CARC{$^\downarrow$}
& DRD2{$^\uparrow$} & hERG{$^\downarrow$} & HIA{$^\uparrow$}
& LIV{$^\downarrow$}
& MUT$^\downarrow$ & PlogP$^\uparrow$ & QED$^\uparrow$
\\
& ($\Delta_p=$)
& 0.1 & 0.1 & 0.2 & 0.1 & 0.2 & 0.1 & 0.1 & 0.1 & 1.0 & 0.1
\\
& ($\Theta_p=$)
& 0.8 & 0.8 & 0.2 & 0.4 & 0.3 & 0.4 & 0.9 & 0.2 & 1.5 & 0.9
\\
\midrule
\multirow{5}{*}{IND}
& \BPQ   
  & –      & \cmark & –      & –      & –      & –      & –      & –      & \cmark & \cmark 
  & 700 & 1,371 
  & 500 & 7
  & \CSO
\\
& \ELQ   
  & –      & –      & –      & –  & \cmark    & – & \cmark & –      & –      & \cmark 
  & 700 
  & 1,376 
  & 500 & 7
  & \GTO
\\
& \ACEP  
  & \cmark & –      & \cmark & –   & \cmark    & –        & –      & –      & \cmark & – 
  & 1,242 & 2,347 
  & 500 & 15
  & \GTO
\\
& \BDPQ  
  & –      & \cmark & –  & \cmark    & –       & –      & –      & –      & \cmark & \cmark
  & 895 & 1,561 %
  & 500 & 13
  & \CSO
\\
& \DHMQ  
  & –      & –      & –   & \cmark & –       & \cmark  & –       & \cmark & –      & \cmark
  & 787 & 1,402 
  & 500  & 9
  & \CSO
\\
\midrule
\multirow{5}{*}{OOD}
& \CDE   
  & –      & –      & \cmark  & \cmark & \cmark  & –     & –      & –      & –      & – 
  & 516 & 832 %
  & 500 & 6
  & \CSO
\\
& \ABMP  
  & \cmark & \cmark & –      & –      & –      & –      & –      & \cmark & \cmark & – 
  & 1,500 & 2,809  %
  & 500 & 15
  & \CSO
\\
& \BCMQ  
  & –      & \cmark & \cmark & –      & –      & –      & –      & \cmark & –      & \cmark
  & 1,398 & 2,696 %
  & 500 & 15
  & \CSO
\\
& \BDEQ  
  & –      & \cmark & –     & \cmark & \cmark   & –     & –      & –      & –      & \cmark
  & 603 & 840 %
  & 500 & 11
  & \CSO
\\
& \HLMPQ 
  & –      & –      & –      & –   & –   & \cmark      & \cmark  & \cmark & \cmark & \cmark
  & 1,800 & 3,329 %
  & 500 & 21
  & \GTO
\\

\bottomrule
\end{tabular}

\begin{tablenotes}[normal,flushleft]
\footnotesize
\item 
``{$\mathcal{P}$-Comb}" denotes the combination of $\mathcal{P}$ properties with multiple objectives.
``\#Pairs" and ``\#Mols", 
denote the number of molecule pairs and unique molecules in training, respectively.
``\#Test" and ``\#Tasks" denote the number of test samples 
and multi-property objectives for a specific property combination, respectively.
``Cat" indicates task category.
\cmark indicates properties included in the task;
– indicates properties not involved.
$^\uparrow$ and $^\downarrow$ indicate whether higher or lower scores
of a given property are desirable.
\end{tablenotes}

\end{threeparttable}
\end{small}
\vspace{-15pt}
\end{table*}

\MOptData comprises a total of 28,266 tasks,
with 27,401 tasks optimizing 
a combination of at least 3 properties.
All tasks in \MOptData are systematically curated
by combining subsets of 10 pharmacologically relevant molecular properties:
\textbf{(1) Penalized LogP (PlogP):} representing
solubility, lipophilicity, synthetic accessibility, and ring complexity
-- higher PlogP is typically preferred in drug candidates;
\textbf{(2) Quantitative Estimate of Drug-Likeness (QED):}
assessing overall drug-likeness by incorporating molecular weight, lipophilicity, and hydrogen bonding ability
-- higher QED is desired for better drug-likeness;
\textbf{(3) Parallel Artificial Membrane Permeability Assay (AMP):}
evaluating drug permeability across the cellular membrane
-- higher AMP indicates improved drug absorption;
\textbf{(4) Blood-Brain Barrier Permeability (BBBP):}
representing the ability of a drug to permeate the blood-brain barrier
-- higher BBBP is essential for CNS drugs;
\textbf{(5) human Intestinal Absorption (HIA):}
indicating the ability of a drug to be absorbed through the gastrointestinal tract
-- higher HIA supports effective absorption of orally administered drugs;
\textbf{(6) human Ether-à-go-go Related Gene inhibition (hERG):}
referring to the drug's ability to inhibit the
human ether-à-go-go related gene, which in turn blocks the potassium channel, causing severe cardiac issues
-- lower hERG is necessary to reduce cardiac risks;
\textbf{(7) Carcinogenicity (CARC):}
indicating the potential of a drug to induce cancer
by damaging the genome or disrupting cellular processes
-- lower CARC is desired for safety;
\textbf{(8) Mutagenicity (MUT):}
referring to the likelihood of a drug causing genetic mutations
-- lower MUT scores are preferred to reduce genotoxicity;
\textbf{(9) Drug-induced Liver Injury (LIV):}
representing a drug's potential to induce liver damage (hepatotoxicity)
-- lower DILI is crucial to reduce toxicity;
\textbf{(10) Dopamine Receptor D2 Inhibition (DRD2):}
indicating binding affinity to dopaminergic pathways
-- higher DRD2 scores are desired for antipsychotic drugs
targeting the DRD2 receptor.

We focus on these 10 properties due to their key role in
determining a drug's pharmacokinetic behavior, toxicity risk,
and overall drug-likeness
-- essential factors in real-world lead optimization.
Moreover, these properties are well-studied and typically
considered in existing optimization benchmarks~\cite{gao2022sample,dey2025gellmo}.
For evaluation,
10 representative property combinations (Section~\ref{sec:app:task})
with 119 multi-objective tasks 
are selected and grouped into 51 IND and 68 OOD tasks. (Section~\ref{sec:data:ind_ood}).
These tasks can be divided into 2 categories:
\textbf{(1) General Drug-Likeness and Toxicity (\GTO):}
tasks focused on broadly applicable molecular properties 
relevant for any successful drug candidate, irrespective of the
specific therapeutic endpoint.
\textbf{(2) Context-Specific Objectives (\CSO):}
tasks involving properties that are specific to the
therapeutic end-point, such as DRD2 inhibition or tissue-specific permeability (e.g., BBBP).

\subsection{Constructing Task-Specific Training Pairs}
\label{sec:data:train}
Following Algorithm~\ref{alg:task_construction},
we construct task-specific training pairs $(M_x, M_y)$
from the dataset curated by \cite{chen2021deep},
which contains 256K molecule pairs satisfying the similarity constraint
(i.e., Tanimoto similarity > 0.6).
%
%
Out of these pairs, we select those that
satisfy all \PropImpv property improvement constraints 
(i.e., $(M_x \prec_{\Delta_p} M_y)_{\forall p \in \PropImpv}$)
and all \PropStable property stability constraints
(i.e., $(M_x \cong_{\Delta_p} M_y)_{\forall p \in \PropStable}$)
for each task optimizing sub-optimal \PropImpv properties
and near-optimal \PropStable properties (Appendix~\ref{sec:app:task:algo}).
For a given task with $\mathcal{P}$ properties,
each property $p \in \mathcal{P}$ is considered 
sub-optimal or near-optimal
based on $\Theta_p$ (shown in Table~\ref{tbl:task_summary}) as
described earlier in Section~\ref{sec:data}.
These thresholds are set to the 60th percentile of all training molecules among 256K pairs,
reflecting desirable scores for an optimized lead molecule.

\subsection{Constructing Task-Specific Test Set}
\label{sec:data:test}

We construct a test set by randomly sampling 250K molecules from
ZINC~\cite{sterling2015zinc},
a widely used subset of commercially available molecules.
All sampled molecules satisfy Lipsinki's rule of 5~\cite{Lipinski2001},
and do not overlap with the training set to ensure no data leakage.
This creates an initial pool of drug-like molecules
having some near-optimal properties with desirable scores,
and some sub-optimal ones requiring further improvement.
From this pool, we select a molecule $M_x$
into the test set of a task improving \PropImpv and 
maintaining \PropStable properties,
if $M_x$ has every property $p\in\PropImpv$ worse than $\Theta_p$,
and every property $p\in\PropStable$ exceeding $\Theta_p$.
This selection ensures a representative test set
for evaluation on diverse multi-objective tasks, given a
specific property combination.
Following this selection process,
we randomly sample 500 molecules for each of
10 representative property combinations in evaluation.
%

\subsection{Quality Control}
\label{sec:data:quality}
We implement several quality control measures, detailed in Appendix~\ref{sec:app:quality},
to ensure the integrity and rigor of \MOptData.
We eliminate duplicate molecules by comparing their canonicalized SMILES representations.
We compute all molecular property scores empirically using established
and widely-used tools
such as ADMET-AI~\cite{swanson2024admet}.
To promote robustness in instruction following, 
we curate 30 distinctly phrased instructions that convey
the same optimization objective using varied semantics
(Appendix~\ref{sec:app:instr}).
To assess LLMs' ability to generalize beyond seen instructions,
we hold out one instruction 
per task during training and use it only during inference.

\subsection{IND and OOD Tasks}
\label{sec:data:ind_ood}
To rigorously evaluate instruction-tuned LLMs on both 
familiar and novel optimization scenarios,
we split the 10 evaluation tasks into 2 groups:
\paragraph{In-Distribution (IND) Tasks:}
IND tasks are defined by property combinations that appear
in the training set.
Performance on these tasks assess
how effectively the model can apply its learned modification strategies
to the exact property combinations and objectives it
was specifically trained on.

\paragraph{Out-of-Distribution (OOD) Tasks:}
OOD tasks involve novel multi-property combinations
and novel multi-property objectives for each combination
that are not used during training (i.e., unseen C-MuMO tasks).
Note that although OOD property combinations are not used in training,
each individual property is still used as part of other combinations in the training tasks.
Success in OOD tasks demonstrates the model's ability to
transfer its knowledge to novel property combinations
and novel multi-objective tasks for each unseen property combination
without task-specific fine-tuning.
This ability is crucial in practice,
where emerging therapeutic goals often necessitate
adapting to previously unseen multi-property trade-offs.

\section{\mollm Models}
\label{sec:model}

We introduce {\mollm}s, a series of general-purpose LLMs
instruction-tuned over \MOptData.
\mollm is tuned to follow property-specific objectives 
in \MOptData.
Instruction tuning over molecule pairs enables 
\mollm to implicitly encode how precise
structural modifications map to multiple property changes~\cite{hansch1969quantitative}.
\mollm learns to apply such targeted modifications to 
improve sub-optimal properties beyond user-defined thresholds
specified in the task instruction.
\mollm also learns to preserve specified
near-optimal properties 
by avoiding structural modifications that would otherwise lower their scores. 
Learning such precise modifications strategies
allows for explicit control over each property with varying objectives.
%

We develop both specialist and generalist {\mollm}s.
Each specialist {\mollm}, denoted as \mollmNTask,
is fine-tuned on a single property combination of $N$ properties,
with multiple objectives in that specific combination.
This enables them to learn focused modification strategies
specific to observed trade-offs for that property combination.
In contrast, generalist {\mollm}s are trained across multiple property combinations and multiple objectives in each combination. 
This promotes knowledge transfer of shared chemical semantics
and modification strategies
to tackle diverse property trade-offs with property-specific objectives.
This enables generalist {\mollm} to act as 
a foundational LLM capable of handling novel tasks
without task-specific retraining,
while offering control over unseen multi-property objectives.

Concretely, we develop a series of generalist {\mollm}s,
denoted as {\mollmNGen}, 
each is jointly trained on multiple C-MuMO tasks involving diverse multi-property, multi-objective combinations with up to $N$ properties.
To train these models,
we fine-tune 2 general-purpose LLMs: 
Mistral-7B-Instruct-v0.3~\cite{mistral2023mistral} and Llama3.1-8B-Instruct~\cite{grattafiori2024llama3herdmodels}
by applying LoRA~\cite{hu2022lora}
on every projection layer and the language modeling head.
We perform 0-shot evaluations (i.e., without in-context examples)
for all {\mollm}s.
For each input molecule, we generate 20 candidates via beam search decoding.
%
Additional details are provided in Appendix~\ref{sec:app:reproducibility}.

\section{Experimental Setup}

\subsection{Baselines}

We compare {\mollm}s against 2 categories of baseline models:
\textbf{(1)} general-purpose LLMs: Mistral-7B Instruct-v0.3~\cite{mistral2023mistral}, Llama-3.1 8B-Instruct~\cite{touvron2023llama},
Claude-3.5 and GPT-4o; and
\textbf{(2)} foundational LLMs for chemistry: 
a Mistral-7B fine-tuned on diverse molecular tasks~\cite{yu2024llasmol},
denoted as \LlaSMolM.
%
Existing non-LLM models
require substantial effort on task-specific tuning or handcrafted reward functions,
making them ill-suited baselines given the scale and diversity of \MOptData.
We use few-shot prompting with only 1 in-context example 
for all general-purpose LLMs
to balance generation quality with computational resources and expenses.
For baselines that support beam-search decoding,
we generate 20 candidate molecules per input using the same generation strategy as in \mollm.
Additional details and prompts are in Appendix~\ref{sec:app:expts_setup:baselines} and Appendix~\ref{sec:app:prompt}, respectively.

\begin{table*}[h!]
\centering
\setlength{\tabcolsep}{0pt}%
\caption{Overall Performance in IND Tasks}
\label{tbl:main_ind}
\vspace{-4pt}
\begin{small}
\begin{threeparttable}
\begin{tabular}{
   @{\hspace{0pt}}l@{\hspace{2pt}}
   @{\hspace{2pt}}r@{\hspace{2pt}}
   @{\hspace{2pt}}r@{\hspace{2pt}}
   @{\hspace{2pt}}r@{\hspace{2pt}}
   @{\hspace{4pt}}c@{\hspace{5pt}} 
   @{\hspace{0pt}}r@{\hspace{2pt}}
   @{\hspace{2pt}}r@{\hspace{2pt}}
   @{\hspace{2pt}}r@{\hspace{2pt}}
   @{\hspace{4pt}}c@{\hspace{5pt}} 
   @{\hspace{0pt}}r@{\hspace{2pt}}
   @{\hspace{2pt}}r@{\hspace{2pt}}
   @{\hspace{2pt}}r@{\hspace{2pt}}
   @{\hspace{4pt}}c@{\hspace{5pt}} 
   @{\hspace{0pt}}r@{\hspace{2pt}}
   @{\hspace{2pt}}r@{\hspace{2pt}}
   @{\hspace{2pt}}r@{\hspace{2pt}}
   @{\hspace{4pt}}c@{\hspace{5pt}} 
   @{\hspace{0pt}}r@{\hspace{2pt}}
   @{\hspace{2pt}}r@{\hspace{2pt}}
   @{\hspace{2pt}}r@{\hspace{0pt}}
}
\toprule
\multirow{2}{*}{Model} & \multicolumn{3}{c}{\BPQ} && \multicolumn{3}{c}{\ELQ} && \multicolumn{3}{c}{\ACEP}
&& \multicolumn{3}{c}{\BDPQ}  && \multicolumn{3}{c}{\DHMQ} \\
\cmidrule(){2-4} \cmidrule(){6-8} \cmidrule(){10-12} \cmidrule(){14-16} \cmidrule(){18-20}
& \SR$^{\uparrow}$ & \Sim$^{\uparrow}$ & \RI$^{\uparrow}$ & 
& \SR$^{\uparrow}$ & \Sim$^{\uparrow}$ & \RI$^{\uparrow}$ &
& \SR$^{\uparrow}$ & \Sim$^{\uparrow}$ & \RI$^{\uparrow}$ &
& \SR$^{\uparrow}$ & \Sim$^{\uparrow}$ & \RI$^{\uparrow}$ &
& \SR$^{\uparrow}$ & \Sim$^{\uparrow}$ & \RI$^{\uparrow}$ \\
\midrule

\rowcolor{lightgray}
\multicolumn{20}{c}{\textbf{General-purpose LLMs}} 
\\
Mistral (0-shot) & 28.80 & \textbf{\underline{0.75}} & 1.24 &  & 21.60 & 0.72 & 0.16 &  & 26.20 & 0.75 & 1.10 &  & 2.40 & \textbf{\underline{0.72}} & 0.49 &  & 4.80 & 0.71 & 0.76 \\
Llama (0-shot) & 33.60 & 0.70 & 0.78 &  & 16.60 & \textbf{\underline{0.74}} & 0.10 &  & 17.20 & 0.74 & 0.69 &  & 8.80 & \textbf{\underline{0.72}} & 1.67 &  & 6.00 & \textbf{\underline{0.73}} & 1.35 \\
Claude-3.5 (0-shot) & 51.80 & 0.68 & 0.89 &  & 20.00 & 0.64 & 0.20 &  & 29.60 & 0.71 & 0.69 &  & 11.20 & 0.67 & 1.80 &  & 5.20 & 0.63 & 1.84 \\
GPT-4o (0-shot) & 30.20 & 0.72 & 0.55 &  & 16.60 & 0.72 & 0.10 &  & 22.20 & 0.74 & 0.52 &  & 4.20 & \textbf{\underline{0.72}} & 3.98 &  & 5.80 & 0.72 & 0.88 \\
\cellcolor{yellow!20}Mistral (1-shot) & 72.80 & 0.63 & 1.26 &  & 74.80 & 0.59 & \underline{0.28} &  & 63.80 & 0.64 & 1.03 &  & 21.60 & 0.59 & \underline{4.76} &  & \underline{\cellcolor{yellow!20}25.60} & \cellcolor{yellow!20}0.55 & \cellcolor{yellow!20}1.89 \\
Llama (1-shot) & 49.60 & 0.68 & 0.95 &  & 36.80 & 0.68 & 0.15 &  & 40.20 & 0.70 & 1.12 &  & 14.40 & 0.63 & 2.65 &  & 13.80 & 0.56 & \textbf{\underline{3.39}} \\
Claude-3.5 (1-shot) & 61.80 & 0.65 & \underline{1.31} &  & 29.20 & 0.63 & 0.21 &  & 32.60 & 0.71 & \underline{1.24} &  & 15.60 & 0.58 & 3.99 &  & 8.40 & 0.65 & 1.38 \\
GPT-4o (1-shot) & 28.60 & 0.74 & 0.77 &  & 19.60 & 0.72 & 0.12 &  & 23.00 & \textbf{\underline{0.76}} & 1.09 &  & 5.60 & 0.68 & 3.47 &  & 5.60 & 0.71 & 1.22 \\

\rowcolor{lightgray}
\multicolumn{20}{c}{\textbf{Foundational LLMs for Chemistry}}
\\
\cellcolor{yellow!20}LlaSMol-M & \underline{\cellcolor{yellow!20}78.20} & \cellcolor{yellow!20}0.64 & \cellcolor{yellow!20}0.92 &  & \underline{\cellcolor{yellow!20}81.40} & \cellcolor{yellow!20}0.62 & \underline{\cellcolor{yellow!20}0.28} &  & \underline{\cellcolor{yellow!20}68.60} & \cellcolor{yellow!20}0.66 & \cellcolor{yellow!20}1.00 &  & \underline{\cellcolor{yellow!20}22.60} & \cellcolor{yellow!20}0.68 & \cellcolor{yellow!20}2.22 &  & 24.80 & 0.62 & 1.44 \\

\rowcolor{lightgray}
\multicolumn{20}{c}{\textbf{Specialist LLMs}}
\\

\cellcolor{green!10}\mollmTaskM & 71.00 & 0.57 & 2.59 &
& 81.80 & 0.55 & 0.39 &
& 85.60 & 0.54 & \textbf{2.46} &
& \textbf{\cellcolor{green!10}56.60} & \cellcolor{green!10}0.50 & \cellcolor{green!10}5.48 &
& \cellcolor{green!10}44.60 & \cellcolor{green!10}0.57 & \cellcolor{green!10}2.96 \\

\cellcolor{green!10}\mollmTaskL 
& \cellcolor{green!10}84.20 & \cellcolor{green!10}0.58 & \cellcolor{green!10}2.09 &  
& \cellcolor{green!10}85.40 & \cellcolor{green!10}0.53 & \textbf{\cellcolor{green!10}0.41} &
& \cellcolor{green!10}88.00 & \cellcolor{green!10}0.54 & \cellcolor{green!10}2.24 &
& 43.60 & 0.58 & 4.85 &
& 35.40 & 0.65 & 2.63 \\

\hline

\ImpT (\%) & 7.7 & -9.4 & 127.2 &  & 4.9 & -14.5 & 46.4 &  & 28.3 & -18.2 & 124.0 &  & 150.4 & -26.5 & 146.8 &  & 74.2 & 3.6 & 56.6 \\

\rowcolor{lightgray}
\multicolumn{20}{c}{\textbf{Generalist LLMs}}
\\

\cellcolor{blue!10}\mollmNGenM
& 84.80 & 0.63 & 2.64 &  
& 83.20 & 0.63 & 0.33 & 
& 86.60 & 0.60 & 2.34 & 
& 50.60 & 0.58 & 4.93 &
& \textbf{\cellcolor{blue!10}53.40} & \cellcolor{blue!10}0.59 & \cellcolor{blue!10}3.26
\\

\cellcolor{blue!10}\mollmNGenL 
& 88.80 & 0.62 & 2.16 &  
& \textbf{\cellcolor{blue!10}90.80} & \cellcolor{blue!10}0.63 & \cellcolor{blue!10}0.34 &  
& \textbf{\cellcolor{blue!10}92.80} & \cellcolor{blue!10}0.58 & \cellcolor{blue!10}2.22 &  
& \cellcolor{blue!10}51.00 & \cellcolor{blue!10}0.58 & \cellcolor{blue!10}5.40 &
& 50.40 & 0.59 & 3.28 
\\

\cellcolor{blue!10}\mollmDecGenM & \textbf{\cellcolor{blue!10}89.40} & \cellcolor{blue!10}0.62 & \cellcolor{blue!10}2.30 &  & 88.40 & 0.59 & \textbf{0.41} &  & 74.60 & 0.61 & 1.92 &  & 48.40 & 0.58 & 5.05 &  & 52.20 & 0.61 & 2.24 \\

\mollmDecGenL & 79.40 & 0.57 & \textbf{2.67} &  & 79.00 & 0.56 & \textbf{0.41} &  & 72.60 & 0.57 & 2.27 &  & 42.60 & 0.55 & \textbf{5.89} &  & 41.80 & 0.57 & 3.32 \\

\hline
\ImpG (\%) & 14.3 & -3.1 & 150.0 &  & 11.5 & 1.6 & 21.4 &  & 35.3 & -12.1 & 122.0 &  & 125.7 & -14.7 & 143.2 &  & 108.6 & 7.3 & 72.5 \\

\bottomrule
\end{tabular}

\begin{tablenotes}[normal,flushleft]
\footnotesize
\setlength{\fboxsep}{1pt}
\item $^\uparrow$ and $^\downarrow$ indicate whether a higher or lower metric is preferred, respectively.
For each task, the best-performing model is in \textbf{bold}, and
the best baseline is \underline{underlined}.
{\ImpT} and {\ImpG} represent the percentage improvement from the \colorbox{green!10}{best specialist LLM} 
and \colorbox{blue!10}{best generalist LLM} over the 
\colorbox{yellow!20}{best baseline},
respectively.
The best model in each group is selected based on {\SR} for each task.
\end{tablenotes}

\end{threeparttable}
\end{small}
\vspace{-14pt}
\end{table*}

\subsection{Evaluation Metrics}

We employ multiple evaluation metrics (detailed in Appendix~\ref{sec:app:eval}) 
to enable a comprehensive assessment. 
For clarity and brevity, we report results primarily using the following metrics:
\textbf{(1) Success Rate} (\SR): 
the proportion of input molecules successfully optimized,
such that all sub-optimal properties are improved, 
and all near-optimal ones are maintained within their corresponding $\Delta_p$
-- reflecting the model's ability to follow property-specific objectives;
%
%
\textbf{(2) Similarity with input} (\Sim): the average Tanimoto similarity~\cite{Bajusz2015why} between the optimized and corresponding input molecule; 
\textbf{(3) Relative Improvement} (\RI): the relative improvement averaged across all sub-optimal properties.
Higher \SR, \Sim, and {\RI} are preferred, denoting more successful and effective optimizations.
In Appendix~\ref{sec:app:results},
we report results with a stricter notion of success, via \SSR,
measuring success only if each property in the task exceeds 
$\Theta_p$. 

\section{Experimental Results}
\label{sec:results}

\paragraph{Main Findings:}
The key findings are summarized as:
\textbf{(1)} Both specialist and generalist {\mollm}s consistently
surpass general-purpose LLMs and foundational LLMs for chemistry
across all IND (Section~\ref{sec:results:ind})
and OOD tasks (Section~\ref{sec:results:ood}),
achieving up to 126\% higher {\SR} and 143\% higher {\RI}.
\textbf{(2)} Generalist {\mollm}s outperform specialist {\mollm}s
on 4 out of 5 IND combinations, with 26\% more successful optimizations on challenging tasks, such as \DHMQ
(Section~\ref{sec:results:ind}).
\textbf{(3)} Generalist {\mollm}s demonstrate remarkable 0-shot
generalization to OOD tasks, surpassing the best general-purpose
LLMs by 35\% in {\SR} and 76\% in {\RI} (Section~\ref{sec:results:ood}).
\textbf{(4)} 
Generalist {\mollm}s exhibit strong generalization 
when prompted with unseen instructions across all IND tasks (Section~\ref{sec:results:uninst}).

\subsection{IND Tasks}
\label{sec:results:ind}

Table~\ref{tbl:main_ind} presents the performance comparison of {\mollm}s
and baselines across all IND tasks.
Detailed task-specific results are in Appendix~\ref{sec:app:results:ind}.

\paragraph{Overall Comparison:}
Across all IND tasks, all specialist and generalist {\mollm}s
consistently outperform all baselines.
Notably, the generalist \mollmDecGenM
outperforms the best baseline by 37\% and 102\% in {\SR} and {\RI} on average,
indicating its superior ability as a foundational model 
to perform targeted modification across diverse C-MuMO tasks.
On two challenging tasks, \BDPQ and \DHMQ,
with a specific therapeutic endpoint (DRD2 inhibition),
both specialist and generalist {\mollm}s
successfully optimize as much as 150\% and 126\% more input molecules
than the baselines, with even 1-fold better {\RI}.
Such strong performance demonstrates the ability of {\mollm}s
to tackle complex property trade-offs.
%

Furthermore, when evaluated under the stricter success criteria (via \SSR)
-- which requires each property to exceed pharmaceutically relevant thresholds (i.e., $\Theta_p$) --
the performance gap between {\mollm}s and baselines becomes even more pronounced.
Table~\ref{tbl:main_ind_strict} demonstrates that generalist {\mollm}s 
outperform the best baseline by as much as 218\% in {\SR} and 313\% in \RI.
This highlights the ability of {\mollm}s to not only optimize more molecules, 
but also to improve each desired property up to significant levels.
%

\vspace{-5pt}
\paragraph{Comparison between specialist and generalist \mollm:}
Table~\ref{tbl:main_ind} demonstrates that 
generalist {\mollm}s outperform specialist ones on 4 out of 5 IND combinations,
with particularly large gains on the challenging \DHMQ tasks.
This trend is prominent in tasks with fewer task-specific 
training pairs, such as \BPQ, \ELQ, and \DHMQ,
where generalist models
outperform specialist ones by up to 26\% in {\SR}.
Limited training pairs in these tasks
hinder the specialist models to learn robust modification strategies.
In contrast, generalist ones benefit from transferable knowledge of property
trade-offs and learn optimization strategies from other diverse multi-property, multi-objective training tasks.

Interestingly, in the \BDPQ tasks, despite having only 895 pairs, \mollmTaskM outperforms all generalist ones.
The generalist variant, \mollmNGen, -- trained only on tasks involving BBBP, DRD2, PlogP and QED -- remains competitive due to its focused training on these specific properties.
In contrast, {\mollmDecGen} -- trained on all possible property combinations 
involving up to 10 properties --
performs worse than \mollmNGen and specialist \mollm.
This could be due to \mollmDecGen encountering tasks with competing or conflicting objectives,
which weakens its ability to specialize in \BDPQ-specific trade-offs.
This highlights a key challenge in developing foundational models:
while multi-task tuning promotes cross-task knowledge transfer,
it may also introduce conflicts that negatively impact 
performance on specialized tasks (e.g., \BDPQ).

\begin{table*}[h!]
\centering
\setlength{\tabcolsep}{0pt}%
\caption{Overall Performance in OOD Tasks}
\label{tbl:main_ood}
\vspace{-5pt}
\begin{small}
\begin{threeparttable}
\begin{tabular}{
   @{\hspace{0pt}}l@{\hspace{5pt}}
   @{\hspace{2pt}}r@{\hspace{2pt}}
   @{\hspace{2pt}}r@{\hspace{2pt}}
   @{\hspace{2pt}}r@{\hspace{2pt}}
   @{\hspace{5pt}}c@{\hspace{5pt}} 
   @{\hspace{0pt}}r@{\hspace{2pt}}
   @{\hspace{2pt}}r@{\hspace{2pt}}
   @{\hspace{2pt}}r@{\hspace{2pt}}
   @{\hspace{5pt}}c@{\hspace{5pt}} 
   @{\hspace{0pt}}r@{\hspace{2pt}}
   @{\hspace{2pt}}r@{\hspace{2pt}}
   @{\hspace{2pt}}r@{\hspace{2pt}}
   @{\hspace{5pt}}c@{\hspace{5pt}} 
   @{\hspace{0pt}}r@{\hspace{2pt}}
   @{\hspace{2pt}}r@{\hspace{2pt}}
   @{\hspace{2pt}}r@{\hspace{2pt}}
   @{\hspace{5pt}}c@{\hspace{5pt}} 
   @{\hspace{0pt}}r@{\hspace{2pt}}
   @{\hspace{2pt}}r@{\hspace{2pt}}
   @{\hspace{2pt}}r@{\hspace{0pt}}
}
\toprule
\multirow{2}{*}{Model} & \multicolumn{3}{c}{\CDE} && \multicolumn{3}{c}{\ABMP} && \multicolumn{3}{c}{\BCMQ} && \multicolumn{3}{c}{\BDEQ} && \multicolumn{3}{c}{\HLMPQ} \\
\cmidrule(){2-4} \cmidrule(){6-8} \cmidrule(){10-12} \cmidrule(){14-16} \cmidrule(){18-20}
& \SR$^{\uparrow}$ & \Sim$^{\uparrow}$ & \RI$^{\uparrow}$ & 
& \SR$^{\uparrow}$ & \Sim$^{\uparrow}$ & \RI$^{\uparrow}$ &
& \SR$^{\uparrow}$ & \Sim$^{\uparrow}$ & \RI$^{\uparrow}$ &
& \SR$^{\uparrow}$ & \Sim$^{\uparrow}$ & \RI$^{\uparrow}$ &
& \SR$^{\uparrow}$ & \Sim$^{\uparrow}$ & \RI$^{\uparrow}$ \\
\midrule

\rowcolor{lightgray}
\multicolumn{20}{c}{\textbf{General-purpose LLMs}} 
\\
Mistral (0-shot) & 3.00 & 0.73 & 1.33 &  & 23.00 & \textbf{\underline{0.77}} & 0.93 &  & 25.40 & 0.69 & 0.25 &  & 3.00 & \textbf{\underline{0.71}} & 1.05 &  & 11.60 & \textbf{\underline{0.79}} & \textbf{\underline{1.76}} \\
Llama (0-shot) & 6.80 & 0.68 & 0.77 &  & 44.60 & 0.71 & 0.61 &  & 20.40 & 0.72 & 0.20 &  & 2.20 & 0.68 & 0.60 &  & 20.20 & 0.72 & 0.68 \\
Claude-3.5 (0-shot) & 6.80 & 0.70 & 1.07 &  & 43.60 & 0.70 & 0.80 &  & 30.00 & 0.64 & 0.26 &  & 4.80 & 0.62 & 0.57 &  & 21.00 & 0.66 & 0.59 \\
GPT-4o (0-shot) & 3.80 & \textbf{\underline{0.74}} & 1.56 &  & 27.00 & 0.73 & 0.51 &  & 19.60 & 0.72 & 0.19 &  & 3.40 & \textbf{\underline{0.71}} & 0.42 &  & 12.80 & 0.72 & 0.47 \\
\cellcolor{yellow!20}Mistral (1-shot) & \underline{\cellcolor{yellow!20}30.60} & \cellcolor{yellow!20}0.62 & \textbf{\underline{\cellcolor{yellow!20}1.66}} &  & \underline{\cellcolor{yellow!20}73.20} & \cellcolor{yellow!20}0.64 & \underline{\cellcolor{yellow!20}1.09} &  & 63.80 & 0.60 & \underline{0.31} &  & \underline{\cellcolor{yellow!20}21.60} & \cellcolor{yellow!20}0.58 & \cellcolor{yellow!20}1.16 &  & \underline{\cellcolor{yellow!20}55.60} & \cellcolor{yellow!20}0.62 & \cellcolor{yellow!20}0.77 \\
Llama (1-shot) & 18.20 & 0.55 & 1.51 &  & 60.80 & 0.70 & 0.83 &  & 41.60 & 0.67 & 0.23 &  & 11.40 & 0.51 & \textbf{\underline{1.54}} &  & 28.00 & 0.70 & 0.75 \\
Claude-3.5 (1-shot) & 8.40 & 0.66 & 1.09 &  & 45.20 & 0.64 & 0.87 &  & 32.40 & 0.61 & 0.30 &  & 7.20 & 0.55 & 1.22 &  & 25.00 & 0.61 & 0.72 \\
GPT-4o (1-shot) & 7.00 & 0.72 & 1.04 &  & 34.40 & 0.74 & 0.65 &  & 23.40 & \textbf{\underline{0.73}} & 0.21 &  & 2.20 & 0.70 & 0.83 &  & 13.40 & 0.71 & 0.65 \\

\rowcolor{lightgray}
\multicolumn{20}{c}{\textbf{Foundational LLMs for Chemistry}}
\\
\cellcolor{yellow!20}\LlaSMolM & 29.80 & 0.61 & 1.28 &  & 72.40 & 0.67 & 0.78 &  & \underline{\cellcolor{yellow!20}72.80} & \cellcolor{yellow!20}0.63 & \cellcolor{yellow!20}0.30 &  & 18.20 & 0.60 & 0.65 &  & 37.80 & 0.68 & 0.66 \\

\rowcolor{lightgray}
\multicolumn{20}{c}{\textbf{Generalist LLMs}}
\\

\cellcolor{blue!10}\mollmDecGenM & \textbf{\cellcolor{blue!10}39.80} & \cellcolor{blue!10}0.58 & \textbf{\cellcolor{blue!10}1.66} &  & \textbf{\cellcolor{blue!10}86.60} & \cellcolor{blue!10}0.63 & \cellcolor{blue!10}1.68 &  & \textbf{\cellcolor{blue!10}84.20} & \cellcolor{blue!10}0.62 & \cellcolor{blue!10}0.42 &  & \textbf{\cellcolor{blue!10}29.20} & \cellcolor{blue!10}0.60 & \cellcolor{blue!10}1.22 &  & \textbf{\cellcolor{blue!10}74.60} & \cellcolor{blue!10}0.61 & \cellcolor{blue!10}1.36 \\

\mollmDecGenL & 33.20 & 0.55 & 1.50 &  & 79.60 & 0.58 & \textbf{1.81} &  & 80.00 & 0.57 & \textbf{0.44} &  & 28.40 & 0.58 & 0.88 &  & 65.40 & 0.58 & 1.35 \\

\hline
\ImpG (\%)
& 30.1 & -6.5 & 0.0 &  & 18.3 & -1.6 & 54.1 &  & 15.7 & -1.6 & 40.0 &  & 35.2 & 3.4 & 5.2 &  & 34.2 & -1.6 & 76.6 \\

\bottomrule
\end{tabular}

\begin{tablenotes}[normal,flushleft]
\footnotesize
\item The metrics, notations and formatting have the same meanings as those
in Table~\ref{tbl:main_ind}.
\par
\end{tablenotes}

\end{threeparttable}
\end{small}
\vspace{-15pt}
\end{table*}

\vspace{-5pt}
\paragraph{Comparison with general-purpose LLMs:}
Table~\ref{tbl:main_ind} shows that all {\mollm}s
consistently outperform all general-purpose LLMs
across all IND tasks,
achieving up to 109\% higher {\SR}
than the best general-purpose LLM, Mistral (1-shot).
This strong performance gap underscores the benefit of instruction tuning
on molecule pairs, which enables {\mollm}s
to learn robust and effective modification strategies
that are difficult for general-purpose LLMs to learn through in-context examples alone.
Moreover, general-purpose LLMs exhibit lower {\RI}
among the limited successfully optimized molecules, compared to {\mollm}s.
This demonstrates the ability of {\mollm}s to perform more targeted
modifications to yield substantial improvements on each sub-optimal property.

\paragraph{Comparison with foundational LLMs for chemistry:}
All {\mollm}s substantially outperform the SoTA
foundational LLM for chemistry, \LlaSMolM, on all IND tasks.
Another foundational LLM, \ChemDFM, performs worse than \LlaSMol (Appendix~\ref{sec:app:results}).
Notably, on \BDPQ and \DHMQ,
\mollmDecGenM achieves a 126\% and 115\% higher {\SR}, respectively,
with higher {\RI}
by 143\% and 126\%, respectively, compared to \LlaSMolM.
While \LlaSMol is instruction-tuned on a broad range of molecular tasks,
{\mollm}s are specifically instruction-tuned on different multi-property optimization tasks.
This highlights the efficacy of instruction-tuning on optimization tasks
to learn targeted modifications and navigate multi-property trade-offs.
Appendix~\ref{sec:app:case_study} presents 2 cases of such targeted modifications.
%

\subsection{OOD Tasks}
\label{sec:results:ood}

Table~\ref{tbl:main_ood} presents the performance of
{\mollm}s and baselines across all OOD tasks.
%
Since {\mollmNTask}s and {\mollmNGen} models use task-specific pairs,
they are inapplicable to OOD tasks.
Overall, generalist {\mollm}s exhibit strong 0-shot generalization to 
novel C-MuMO tasks,
consistently outperforming all baselines.
Specifically, the best-performing generalist model, \mollmDecGenM,
achieves an average {\SR} of 63\% across all tasks,
outperforming the best baseline, Mistral (1-shot),
by as much as 35\% and 77\% in {\SR} and {\RI}, respectively.
These strong results demonstrate the remarkable ability of generalist {\mollm}s to learn transferable optimization strategies 
and tackle unseen controllable property-specific objectives during inference.  
Such generalizability is crucial in practice, 
where evolving therapeutic goals often introduce novel property combinations and novel objectives.

\subsection{Generalizability to Unseen Instructions}
\label{sec:results:uninst}

Table~\ref{tbl:main_uninst} compares specialist {\mollm}s
with generalist {\mollm}s when evaluated with a hold-out instruction
and property name (Appendix~\ref{sec:app:instr}).
Overall, specialist {\mollm}s exhibit a performance
drop of over 5\% in {\SR} on 2 out of 5 IND combinations.
In contrast, generalist {\mollm}s retain consistent
performance on all tasks.
This indicates that generalist models --
trained on more tasks and instructions --
can generalize better to unseen instructions with different phrasings.
Such generalizability is crucial in practice, where
task instructions can vary widely.
Notably, {\mollmDecGenL} demonstrates more robustness than
{\mollmDecGenM}, 
reflecting a reduced tendency to overfit to specific wordings.

\vspace{-5pt}
\section{Conclusion}
\vspace{-5pt}
In this paper, we introduced \MOptData, 
the first instruction-tuning dataset enabling controllable 
molecule optimization with property-specific objectives. 
Leveraging \MOptData, we developed {\mollm}s,
that consistently and largely outperform
strong general-purpose LLMs and foundational LLMs for chemistry 
across all IND and OOD tasks. 
Moreover, generalist {\mollm}s exhibit strong generalization 
to unseen tasks, outperforming baselines by 27\% on average. 
This indicates the potential of {\mollm} as a foundational model 
to tackle diverse tasks with realistic, controllable objectives reflecting real-world scenarios.


\section{Limitations}
\label{sec:limitations}

While our work represents a significant step toward controllable, multi-objective molecule optimization, several limitations remain:
\textbf{(1)} Our current framework is designed for single-step optimization. 
In practice, optimizing molecules to reach pharmaceutically meaningful thresholds for all properties 
may require multiple iterative modifications. 
Designing a feedback mechanism for \mollm 
or intermediate reward signal to guide iterative refinement 
is non-trivial and is a direction for future work.
\textbf{(2)} We rely on computational predictors for molecular properties. 
Although they are well-established and widely used, 
they may introduce inaccuracies and may not always reflect exact experimental outcomes. 
Incorporating experimentally validated datasets or
feedback to LLMs with wet-lab data is a promising direction for future work.
\textbf{(3)} Although we demonstrate strong generalization to unseen instructions, our instruction templates are still synthetically generated. 
Future work could explore more diverse linguistic variation to test LLM robustness in truly open-ended settings.

\section{Impact Statement}
\label{sec:impact}

This work presents the first instruction-tuning dataset, \MOptData, that explicitly supports property-specific objectives in multi-property molecule optimization
-- enabling models to selectively improve sub-optimal properties while preserving near-optimal ones.
Built on this dataset, our developed instruction-tuned LLMs (\mollm) represent a substantial advancement toward 
controllable molecule optimization, addressing practical drug design requirements often overlooked by existing approaches.
{\mollm}s consistently outperform both strong general-purpose LLMs and 
foundational LLMs for chemistry across 
challenging optimization tasks involving conflicting objectives.
By demonstrating robust generalization to novel property combinations 
and novel multi-property constraints, 
\mollm paves the way for scalable, general-purpose foundation LLMs that can flexibly handle diverse drug design constraints.
We anticipate that {\mollm}
will serve as a building block for future iterative LLM optimization frameworks.

\paragraph{Broader Impacts:}
%
The development of foundational LLMs for controllable multi-property 
molecule optimization represents a 
significant step toward AI-based molecular design tools. 
Their ability to follow property-specific instructions enables iterative optimization workflows, 
where molecules are refined over multiple steps based on intermediate feedback 
-- 
a common and necessary paradigm in real-world lead optimization. 
Through natural language instructions, 
these models can be flexibly adapted to a variety of drug design scenarios without extensive retraining.
Such flexibility lowers the barrier to deploying intelligent drug design pipelines, especially for researchers with limited computational or domain resources. 
Ultimately, such scalable and generalizable frameworks have the potential to accelerate early-stage drug development,
reduce experimental burden, 
and democratize access to advanced drug design capabilities.

\section{Ethics Statement}
\label{sec:ethics}

Our work introduces instruction-tuning dataset, \MOptData
and {\mollm}s tuned on \MOptData 
for multi-property molecule optimization. 
While \MOptData is curated with drug-like molecule
and to improve pharmaceutically relevant and desirable properties, 
we cannot fully guarantee the absence of harmful compounds or the potential for misuse.
Notably, 4 of the 10 properties in {\MOptData} 
-- carcinogenicity, hERG inhibition, drug-induced liver injury, and mutagenicity
-- are directly related to drug toxicity. 
Our models are explicitly tuned to minimize these property scores, 
and thus, to improve drug safety profiles 
aligned with widely accepted pharmacological desirability. 
The objective is to generate drug-like molecules with reduced toxicity, not to increase toxicity or discover harmful compounds.

Given that our models are fine-tuned on 
general-purpose open-source LLMs, 
they may still retain knowledge about toxic substructures or chemicals from the broader pretraining corpus. 
While our instruction-tuning encourages models to generate molecules
with more pharmaceutically desirable profiles, 
we cannot fully eliminate the possibility of generating undesirable molecules if misused or prompted adversarially.

We strongly discourage any application of {\mollm}s outside responsible drug discovery research. 
Deployment of these models should be accompanied by toxicity screening, expert review, and strong usage controls. 
We expect all users of our dataset and models to uphold the highest standards of ethical research and to take appropriate precautions to prevent unintended consequences.

\bibliography{paper}

\clearpage

\appendix

\renewcommand{\thefigure}{A\arabic{figure}} 
\setcounter{figure}{0} 

\renewcommand{\thetable}{A\arabic{table}} 
\setcounter{table}{0} 

\setcounter{algocf}{0}
\renewcommand{\thealgocf}{A\arabic{algocf}}

\section{Related Work}
\label{sec:related}
Computational approaches 
have primarily focused on single- or double-property optimization tasks~\cite{you2018graph, blaschke2020reinvent, xie2021mars, bung2022silico,sun2022molsearch}. 
Graph-based methods such as Modof~\cite{chen2021deep}, 
MIMOSA~\cite{fu2021mimosa},
and f-RAG~\cite{lee2024molecule} 
perform substructure modifications on molecular graphs, 
while sequence-based methods 
like Chemformer~\cite{irwin2022chemformer} and Prompt-MolOpt~\cite{wu2024leveraging},
formulate optimization as translation tasks over SMILES strings. 
Genetic algorithm-based methods,
GraphGA~\cite{jensen2019graph} and MolLeo~\cite{wang2024efficient}
can optimize multiple properties but
generate entirely new molecular scaffolds, limiting their practical utility.
%
%
Furthermore, existing methods~\cite{jensen2019graph, wang2024efficient, kim2024genetic,yang2021hit},
require task-specific fine-tuning and expert-curated reward functions to model multi-property trade-offs,
limiting their scalability and applicability.

Recently, LLMs have demonstrated great promise for molecule optimization
through natural language instructions~\cite{chang2024}.
ChatDrug~\cite{liu2023chatgpt} and Re3DF~\cite{le2024utilizing}
adopt multi-turn dialogue frameworks for iterative optimization. 
However, their reliance on closed-source APIs leads to high costs.
DrugAssist~\cite{ye2025drugassist} developed
task-specific instruction-tuned LLMs limited to optimization tasks with up to 2 properties.
%
%
%
~\citet{dey2025gellmo} introduced {\OptData} -- 
a large-scale instruction-tuning dataset specifically focused
on multi-property optimization tasks involving 3 or more properties -- 
and further demonstrated the remarkable generalization abilities
of instruction-tuned LLMs. 
However, \OptData does not provide controllable property-specific objectives
required to mimic realistic C-MuMO tasks.
%
%

\section{Details on \MOptData}
\label{sec:app:task}

\subsection{Details on Task Construction}
\label{sec:app:task:algo}


\begin{algorithm*}
\caption{C-MuMO Task Construction from a Molecule Pair}
\label{alg:task_construction}

\KwIn{Molecule pair $(M_x, M_y)$, Pharmaceutically-relevant levels $\{\Theta_p\}$, Improvement thresholds $\{\Delta_p\}$, Set of properties $\mathcal{P}$}

\KwOut{List of valid C-MuMO tasks $\mathcal{T}$ for $(M_x, M_y)$ with at most $\mathcal{P}$ properties}

Initialize $\mathcal{T} \gets \emptyset$ \;

\ForEach{$p \in \mathcal{P}$}{
    Compute $\texttt{change}[p] \gets p(M_y) - p(M_x)$ \;
    Set $\texttt{dir}[p] \gets$ ($\texttt{change}[p] > 0$) if higher $p$ is desirable, else negative \;
}



// \textbf{Identify Sub-optimal and near-optimal Properties:} 

{\PropImpv} $\gets \{p \in \PropImpv \mid \texttt{abs(change)}[p] > \Delta_p\}$ \;
\PropStable $\gets \{p \in \PropStable \mid \texttt{abs(change)}[p] \leq \Delta_p$ and $p(M_x) \succeq \Theta_p\}$ \;

\ForEach{property subset $\mathcal{C} \subseteq \mathcal{P}$ with $|\mathcal{C}| \geq 1$}{
    $\mathcal{C}_i \gets C \cap \PropImpv$ \tcp*[f]{Identify sub-optimal subset} \;
 
    \If{$\mathcal{C}_i = \emptyset$}{\textbf{continue} \tcp*[f]{Skip if no sub-optimal properties} }
    
    \If{not all $\texttt{dir}[p]$ in $\mathcal{C}_i$ are the same}{
        \textbf{continue} \tcp*[f]{Require improvement in all sub-optimal ones}
    }

    \texttt{NeedSwap} $\gets$ true if all $\texttt{dir}[p]$ in $\mathcal{C}_i$ are opposite of desired \tcp*[f]{Determine swap condition} \;

    \If{\texttt{NeedSwap}}{%
        Swap $M_x \leftrightarrow M_y$ \tcp*[f]{Ensure correct direction of improvement} \;
    }
    
    $\mathcal{C}_s \gets C \cap \PropStable$ \tcp*[f]{Identify near-optimal subset} \;
    
    Construct task $t = (M_x, M_y, \mathcal{C}_i, \mathcal{C}_s)$ \tcp*[f]{An optimization task} \;
    
    $\mathcal{T} \gets \mathcal{T} \cup \{t\}$
    
}
\Return{$\mathcal{T}$}
\end{algorithm*}

Algorithm~\ref{alg:task_construction} presents a pseudocode for
constructing all valid C-MuMO tasks with all possible property combinations
involving up to $\mathcal{P}$ properties,
given a molecule pair $(M_x, M_y)$.
To construct \MOptData, we run Algorithm~\ref{alg:task_construction}
on a random sample of 100K molecule pairs sourced from \citet{chen2021deep}.
To create training pairs for a given combination with $N$ properties,
we select only those tasks out of all C-MuMO tasks that have all $N$ properties involved.
For example, 
to create task-specific training pairs for \BDPQ,
we select only tasks that involve all 4 properties:
$\mathcal{T}_{\text{BDPQ}} = \{t=(M_x,M_y,\mathcal{C}_i,\mathcal{C}_s) \in \mathcal{T} \mid (\mathcal{C}_i \cup \mathcal{C}_s) = \mathcal{P}\}$
where $\mathcal{P} = \{$BBBP, DRD2, PlogP and QED$\}$.

We use at most 100 molecule pairs for each C-MuMO task 
(i.e., a unique property combination with explicit property-specific objectives) to balance efficiency and task diversity. 
Given that \MOptData contains over 28K such tasks, 
training a generalist model with all possible pairs would be computationally prohibitive and may overemphasize overrepresented tasks. 
Limiting the number of examples per task ensures that the instruction-tuned model is exposed to a broad spectrum of multi-property trade-offs without biasing toward specific tasks. 
This design supports better generalization across diverse optimization objectives while keeping training tractable.


\subsection{Details on Quality Control}
\label{sec:app:quality}

To ensure a high-quality instruction-tuning dataset,
we applied a series of quality control procedures.

\paragraph{Molecule Deduplication and Canonicalization:}
All molecules in \MOptData are represented using canonical SMILES strings~\cite{Weininger1988smiles}, 
standardized via RDKit~\cite{rdkit}. 
We remove molecules with identical canonicalized SMILES that are structurally equivalent, thereby eliminating redundancy and ensuring that each molecule appears only once. 
%

\paragraph{Empirical Property Computation:}
\MOptData uses computationally predicted scores to annotate each molecule with 10 pharmacologically relevant molecular properties. 
These scores are computed using well-established, high-performing tools widely used in the molecular machine learning community.
Specifically, 
we adopt the official implementation from \citet{you2018graph} for
computing DRD2 and PlogP scores,
and leverage the ADMET-AI tool~\cite{swanson2024admet} to compute all other properties. 
These tools rank among the top-performing predictors in the Therapeutics Data Commons (TDC) benchmark~\cite{Catacutan2024}, and have been extensively validated and adopted in recent studies~\cite{WEI2024,Thomas2024,Wahnou2024,dey2025gellmo,averly2025liddia}.
They provide a reliable, computationally efficient means 
to estimate property scores at scale, enabling the construction of high-quality datasets with broad coverage of chemical space.

While these predictors are not experimentally validated, they demonstrate strong alignment with experimentally measured values and are widely accepted as practical surrogates in virtual screening pipelines. 
Notably, experimentally validated measurements 
are severely limited for many key pharmacological properties. 
For instance, public datasets contain fewer than 2,000 experimentally measured BBBP values -- 
orders of magnitude below what is needed to train large-scale deep learning models or instruction-tuned LLMs. 
Given these constraints, the use of empirical predictors is not only standard but necessary for enabling scalable dataset creation and evaluation.

\paragraph{Instruction Diversity and Generation:}
To avoid LLM overfitting to specific phrasings and 
to promote generalization to natural word variations in task formulation, 
we ensure that each optimization task is associated with a diverse set of instructions. 
Starting from a manually written seed prompt, we use GPT-4o~\cite{openai2024gpt4technicalreport} 
to generate several paraphrased variants that preserve the semantic intent while differing in structure and wording. 
From these, we select 30 semantically equivalent but syntactically diverse instructions per task to include in the training data.

To explicitly assess the models' ability to generalize to new instructions, we hold out one instruction per task as unseen during instruction-tuning. 
This unseen instruction is then used during evaluation to measure robustness to novel phrasings. 
This design allows us to evaluate not only task-level generalization but also linguistic flexibility in following diverse natural language instructions.
All instructions used in training and testing are provided in Appendix~\ref{sec:app:instr}.

\subsection{Details on IND Tasks}
\label{sec:app:task:ind}

\begin{enumerate}[leftmargin=*]

\item
\BPQ (BBBP, PlogP, QED): 
This task involves 7 diverse combinations of property-specific objectives across BBBP, PlogP, and QED
-- three properties central to CNS drug design. 
Each optimization task may involve improving one or more of these properties while maintaining or improving the others.
Optimizing 7 diverse multi-objective combinations of BBBP, PlogP, and QED  simulates early-stage filtering of CNS-active hits. 

\item
\ELQ (hERG, LIV, QED):
Here, the focus is on toxicity-related properties and overall drug-likeness. 
hERG inhibition and liver toxicity are two major causes of clinical trial failures, while QED ensures retained drug-like features. 
A good optimizer must reduce toxicity signals while preserving beneficial characteristics, reflecting real-world needs in late-stage lead optimization, where safety issues are addressed without sacrificing potency.

\item
\ACEP (AMP, CARC, hERG, PlogP):
This task consists of 15 optimization combinations focused on absorption and toxicity-related properties. Each task may require improving any subset of AMP (permeability), CARC (carcinogenicity), hERG (cardiotoxicity), or PlogP (lipophilicity), while stabilizing the rest. It captures the complex trade-offs typical in preclinical candidate refinement, where ADME and safety must be simultaneously addressed.

\item
\BDPQ (BBBP, DRD2, PlogP, QED):
This combination includes 13 challenging optimization tasks 
for antipsychotic drug design.
These require optimization for BBB penetration and DRD2 activity
-- two critical endpoints for efficacy
-- while maintaining lipophilicity and drug-likeness. 
It embodies a highly targeted CNS design task and is one of the most challenging due to strong interdependencies among all properties.

\item
\DHMQ (DRD2, HIA, MUT, QED):
This combination involves optimization of 9 different multi-objective
tasks to optimize a CNS drug target
that must bind to DRD2 receptors while exhibiting high intestinal absorption and low mutagenicity. 
Each task selectively improves or maintains a subset of these properties.
It simulates a realistic challenge in optimizing orally active CNS agents under ADMET and pharmacological constraints.

\end{enumerate}

\subsection{Details on OOD Tasks}
\label{sec:app:task:ood}

\begin{enumerate}[leftmargin=*]

\item \CDE (CARC, DRD2, hERG):
These tasks target CNS drug candidates, 
especially antipsychotics, requiring high DRD2 inhibition. 
However, many such drugs are known to block the hERG potassium channel, 
raising serious cardiotoxicity concerns. 
Additionally, reducing carcinogenicity is essential for long-term drug safety.  Each task may involve increasing DRD2 inhibition while reducing or preserving carcinogenicity and cardiotoxicity.
This mirrors real-world lead optimization, where enhancing efficacy must be carefully balanced against major safety liabilities.

\item \ABMP (AMP, BBBP, MUT, PlogP):
Tasks in this combination target oral CNS-targeted drug design. 
AMP and BBBP capture permeability at intestinal and blood-brain barriers, respectively, essential for drugs acting on the brain after oral administration. 
Mutagenicity must be minimized or maintained to prevent genotoxic effects, while plogP should be improved or maintained to balance lipophilicity, solubility, and synthetic accessibility. 
The task requires coordinated improvement of absorption and brain penetration while constraining safety and physicochemical properties, posing a non-trivial optimization challenge.

\item \BCMQ (BBBP, CARC, MUT, QED):
These tasks comprise 15 multi-objective combinations 
requiring improvements in BBB permeability 
while maintaining or minimizing toxicity (CARC, MUT) and 
retaining or improving drug-likeness (QED). 
Each task emphasizes safety-aware design for CNS-targeting molecules without degrading overall molecular quality.

\item \BDEQ (BBBP, DRD2, hERG, QED):
This combination consists of 11 diverse optimization objectives.
High BBBP and DRD2 inhibition are necessary for efficacy, 
while low hERG inhibition is essential to avoid cardiotoxicity. 
QED must remain high to ensure overall molecular quality. 
This combination embodies the classic efficacy-safety trade-off, 
making it one of the most realistic and challenging multi-objective scenarios.

\item \HLMPQ (HIA, LIV, MUT, PlogP, QED):
This combination includes 21 broad-spectrum ADMET-focused multi-objective tasks aimed at orally administered drugs. 
Each task challenges the model to find precise modifications that jointly optimize oral bioavailability and structural quality while minimizing major toxicity risks
-- reflecting a realistic early-phase development setting.

\end{enumerate}

\section{Diverse Instructions}
\label{sec:app:instr}

Figure~\ref{fig:instr} presents the prompt
template used for instruction-tuning.
Each prompt has three parts: (1) `\{general instruction\}',
(2) input source molecule and properties to adjust for the specific optimization task,
and (3) target optimized molecule.

\begin{figure*}[h!]
\begin{tcolorbox}[
  colback=lightergray, 
  colframe=black, 
  sharp corners, 
  boxrule=0.5pt, 
  width=\textwidth,
  left=1mm, 
  right=1mm, 
  top=1mm, 
  bottom=1mm 
]
\begin{lstlisting}[
  language=, 
  basicstyle=\ttfamily\footnotesize, 
  breaklines=true, 
  breakindent=0pt, % Disable indentation for wrapped lines
  showstringspaces=false, % Remove visible spaces
  xleftmargin=0pt, % Remove left margin
  xrightmargin=0pt, % Remove right margin
  aboveskip=0pt,belowskip=0pt
]
[INST]
{general instruction}

%%% Input : <SMILES> {source-smiles} </SMILES>
%%% Adjust: {adjust_i} {property_i}, ..., {adjust_k} {property_k}
[/INST]

%%% Response: {target-smiles}
\end{lstlisting}
\end{tcolorbox}
\caption{Prompt template used for instruction-tuning {\mollm}s}
\label{fig:instr}
\end{figure*}

The `\{general instruction\}' will be replaced with one of 6
diverse task instructions, which are presented below.
%
The first instruction is manually written, 
and is provided as the seed instruction to GPT-4o
to generate 5 more differently phrased instructions.
The last one is the hold-out instruction for inference.
Below are 6 diverse instructions:

\begin{enumerate}[leftmargin=*]
\item
``Your task is to modify the given molecule to adjust specific molecular properties so that the resulting molecule satisfies the given target thresholds. Keep structural changes as minimal as possible. Your response should only contain a valid SMILES representation of the modified molecule enclosed in <SMILES> </SMILES> tags. The property values of the new molecule should meet or exceed the specified targets enclosed in <THRESHOLD> </THRESHOLD> tags."

\item 
``Adjust the molecular structure to ensure that each specified property reaches the corresponding threshold listed in <THRESHOLD> </THRESHOLD>. Minimize structural changes and try to maintain the core scaffold. Return the resulting molecule using <SMILES> </SMILES> tags."

\item 
``Alter the molecule to satisfy the provided property thresholds in <THRESHOLD> </THRESHOLD>. Preserve the core scaffold and make as few structural changes as possible. Output the SMILES of the new molecule, enclosed in <SMILES> </SMILES>."

\item
``Update the given molecule so that the specified properties fall within acceptable ranges defined by the values in <THRESHOLD> </THRESHOLD>. Maintain as much of the original structure as possible. Output only the modified molecule enclosed in <SMILES> </SMILES> tags."

\item 
``Edit the molecular structure so that all required properties match or exceed the threshold values defined in <THRESHOLD> </THRESHOLD>. Try to retain the core scaffold. Output only the SMILES representation of the optimized molecule enclosed in <SMILES> </SMILES>."

\item 
``Modify the molecule to bring its properties to at least the levels defined in <THRESHOLD> </THRESHOLD>. Avoid excessive modifications and preserve the core scaffold. Output only the resulting molecule's SMILES wrapped in <SMILES> </SMILES>."

\end{enumerate}

In the 2nd part of the prompt template,
multiple properties to be adjusted are described 
via the task-specific `\{adjust\_i\}' (Figure~\ref{fig:instr}).
Each `\{adjust\_i\}' is randomly replaced with one of the following 5 adjustment templates for
each sub-optimal property improvement:
\begin{enumerate}
    \item "{change} {property} to be {direction} <THRESHOLD> {value} </THRESHOLD>",
     \item   "{change} the value of {property} to be {direction} <THRESHOLD> {value} </THRESHOLD>",
      \item  "{change} {property} aiming for {direction} <THRESHOLD> {value} </THRESHOLD>",
      \item  "{change} {property} so it is {direction} <THRESHOLD> {value} </THRESHOLD>",
      \item  "{change} {property} with a goal of {direction} <THRESHOLD> {value} </THRESHOLD>"
\end{enumerate}

Thus, 6 diverse general instruction templates and 5 diverse adjustment templates
together lead to 30 different templates for instruction tuning.

\paragraph{Property Names:}
We used the following names for each property 
where the former is used during instruction-tuning
and the latter is used for evaluation in
the unseen instruction setting.
For other evaluation settings, we used the same
property name as used in tuning.

\begin{enumerate}[leftmargin=*]
    \item AMP: ``membrane permeability", ``Parallel Artificial Membrane Permeability (PAMPA)"
    \item BBBP: ``BBB permeability", ``Blood-brain barrier permeability (BBBP)"
    \item CARC: ``carcinogenicity", 
    ``potential to disrupt cellular metabolic processes"
    \item DRD2: ``DRD2 inhibition", ``inhibition probability of Dopamine receptor D2"'
    \item "hERG": ``hERG inhibition", "potential to block hERG channel",
    \item HIA: ``Intestinal adsorption", ``human intestinal adsorption ability"
    \item "DILI": "liver injury risk", "potential to cause liver disease",
    \item MUT: ``Mutagenicity", ``probability to induce genetic alterations (mutagenicity)"
    \item PlogP: ``Penalized octanol-water partition coefficient (penalized logP)", 
    ``Penalized logP which is logP penalized by synthetic accessibility score and number of large rings"
    \item QED: ``QED", ``drug-likeness quantified by QED score"
\end{enumerate}

\section{Details on Experimental Setup}
\label{sec:app:expts_setup}

\subsection{{\mollm}s}
\label{sec:app:reproducibility}
%
We develop specialist and generalist {\mollm}s by
instruction-tuning general-purpose LLMs on \MOptData using 
specific and multiple property combinations, respectively.
The generalist {\mollmNGen} refers to a generalist model that is trained
on property combinations, each with up to $N$ properties. 
%
%
%
For backbone models, 
we use Mistral-7B-Instruct-v0.3~\cite{mistral2023mistral} 
and Llama3.1-8B-Instruct~\cite{grattafiori2024llama3herdmodels}, 
and apply parameter-efficient fine-tuning using LoRA~\cite{hu2022lora} through the Huggingface Transformers framework~\cite{wolf2020transformers}.
All models are fine-tuned with a learning rate of $1\times10^{-4}$, and a cosine scheduler with 5\% warm-up. 
Specialist models are trained with a batch size of 32 for 10 epochs;
\mollmNGen models are trained with a batch size of 128 for 5 epochs when $N<=4$,
and for 1,800 steps when $N=10$.
The difference in training steps/epochs is to strike a balance between 
training cost and overfitting. 
LoRA is configured with rank 16, $\alpha=16$, dropout rate of 0.05, and is applied to all projection layers and the language modeling head.
We conduct 0-shot evaluation for all {\mollm}s, 
where no in-context examples are provided. 
For each test molecule, we generate 20 candidate molecules 
using beam search decoding with a beam width of 20.

Upon applying LoRA, the number of trainable parameters vary from
42M for Mistral-7B-v0.3 to 44M for Llama3.1-8B-Instruct. 
Training time on a single NVIDIA A100 GPU (40 GB) ranges from 
~1 hour for specialist models to 
8–20 hours for generalist models, 
depending on the total number of tasks and molecule pairs 
-- going up to 28K tasks and 1M pairs for \mollmNGen with N=10. 
The entire training consumed approximately 150 GPU hours.

\subsection{Baselines}
\label{sec:app:expts_setup:baselines}

In this section, we detailed the baselines selected
for our comparison.
Table~\ref{tbl:baseline_LLMs} lists the sources and licenses of all the
source datasets and models (i.e., artifacts) used in this work.
We ensured that all artifacts were utilized in accordance 
with the usage guidelines specified by their original authors or licensors.
For the models we developed, we have considered relevant ethical implications, which are discussed in Section~\ref{sec:ethics}.

\paragraph{General-purpose LLMs:}

We benchmark 4 publicly available general-purpose LLMs,
including 2 open-weights LLMs:
Mistral-7B Instruct-v0.3~\cite{mistral2023mistral},
Llama-3.1 8B-Instruct~\cite{touvron2023llama},
and 2 closed-weights LLMs: Claude-3.5, and GPT-4o
to assess their performance in molecule optimization tasks.
For open-weights LLMs, we utilize their official HuggingFace checkpoints, while for closed-weights ones, 
we access the checkpoints via their official APIs.

We perform 0-shot and 1-shot inference (i.e., with 0 and 1 in-context examples, respectively) using the prompt templates, detailed in Appendix~\ref{sec:app:prompt:glm}.
While few-shot prompting can improve performance, we selected 1-shot as a practical trade-off to control inference cost, especially for closed-sourced API-based models.
Moreover, we found negligible performance improvement using 5-shots in our preliminary experiments.
We generate up to 20 molecules per input molecule using the
same generation strategy for open-source LLMs as in {\mollm}s.
Since Claude and GPT do not support the beam-search decoding strategy
or any customized strategy for multiple sequence generations,
we generate only one molecule per input prompt.
%

\paragraph{Foundational LLMs for Chemistry:}

We adopt \LlaSMolM, the Mistral-7B variant of \LlaSMol, 
as the foundational LLM for chemistry 
due to its strong performance across diverse molecular tasks. 
In comparison to other instruction-tuned LLMs for chemistry, such as 
\ChemDFM~\cite{Zhao2025},
MolInst~\cite{fang2024molinstructions} and ChemLLM~\cite{zhang2024chemllm}, \LlaSMolM consistently achieves state-of-the-art results.
For evaluation, we adopt 0-shot inference. 
Our preliminary experiments indicated that 
incorporating in-context examples did not lead to consistent improvements,
rather impacted performance. 
Furthermore, we employ a simplified prompt format (as shown in Appendix~\ref{sec:app:prompt:llasmol}) 
after observing that \LlaSMol struggles to follow more
complex and structured instruction formats.
For \ChemDFM, we use 0-shot inference using the same prompt template
and generation configuration as of general-purpose LLMs.

\paragraph{Non-LLM Domain-expert Methods:}
Existing non-LLM methods\cite{fu2021mimosa,sun2022molsearch,Angelo2023,kim2024genetic} rely on genetic algorithms or reinforcement learning. 
These methods typically require carefully curated fitness or reward functions to balance multiple properties. 
Such functions are often difficult to design and require significant domain
expertise, limiting their flexibility and generalizability. 

Furthermore, these methods follow a fundamentally different experimental setting:
given an initial pool of candidates,
these methods iteratively modify molecules based on oracle feedback.
This often leads to generating molecules with entirely new scaffolds. 
In contrast, our setting closely aligns with lead optimization in drug discovery,
where the goal is to minimally modify an input molecule while preserving its core scaffold.

\subsection{Evaluation Metrics}
\label{sec:app:eval}

We adopt multiple evaluation metrics to 
comprehensively assess model performance. The metrics are defined as follows:

\begin{enumerate}[leftmargin=*]

\item \textbf{Success Rate (\SR):}
{\SR} denotes the proportion of test cases where at least one of the 20 generated candidate molecules satisfies all specified property objectives
-- i.e., improving all sub-optimal properties while preserving all
near-optimal ones.
When multiple candidates are optimized, 
the molecule exhibiting the highest cumulative improvement 
is selected for evaluation.
A higher {\SR} reflects the model’s effectiveness in achieving task-specific optimization goals.

\item \textbf{Strict Success Rate (\SSR):}
\SSR -- a stricter variant of {\SR} --
measures the proportion of test cases 
where at least one generated molecule not only improves all sub-optimal properties but also brings each of them above the pharmaceutically relevant threshold $\Theta_p$, 
while still preserving all near-optimal properties within their respective $\Delta_p$ bounds.
This metric reflects whether the model can 
generate molecules with desirable properties as specified.

\item \textbf{Validity (\Val):}
Validity refers to the percentage of test instances 
for which at least one of the generated molecules is chemically valid, determined via successful parsing by RDKit.
High \Val ensures the model's ability to generate syntactically correct and chemically valid structures.

\item \textbf{Similarity (\Sim):}
{\Sim} measures the average Tanimoto similarity between optimized and input molecules based on binary Morgan fingerprints (with radius of 2
and dimension of 2048).
Higher {\Sim} indicates better preservation of the similarity constraint
-- 
a key requirement in lead optimization, where maintaining the core molecular scaffold is essential.

\item \textbf{Novelty (\Nov):}
Novelty quantifies the fraction of optimized molecules 
that are not present in the training set.
This indicates the model’s ability to generate novel 
and previously unseen drug candidates, 
crucial for exploration in drug discovery pipelines.

\item \textbf{Synthetic Accessibility Score (\SAS):}
{\SAS} evaluates how easy a molecule is to synthesize, 
with scores ranging from 1 (easily synthesizable) 
to 10 (difficult to synthesize)~\cite{Ertl2009}.
Lower scores indicate simpler, more synthesizable molecules.

\item \textbf{Relative Improvement (\RI):}
{\RI} is computed as the average relative gain 
in each sub-optimal property compared to the input molecule.
This metric reflects the magnitude of property-level improvements achieved by the model.
Formally, for a task improving \PropImpv properties, {\RI} 
is computed as the average of
relative change (\RI$_p$) in each property $p \in \PropImpv$ as:
\begin{equation*}
 \RI = \frac{\sum_{\scriptsize{p \in \PropImpv}} \RI_p}{|\PropImpv|},
\end{equation*}
where \RI$_p$ is computed as:
\begin{equation*}
    \RI_p = \frac{\mathbb{D}[p](p(M_y)-p(M_x))}{p(M_x)},
\end{equation*}
where 
$\mathbb{D}[p]$ is an indicator function denoting whether higher scores of $p$ is desirable,
$p(M_x)$ and $p(M_y)$ denote the score of property $p$ in 
the input molecule $M_x$ and generated molecule $M_y$, respectively.

\item \textbf{Average Property Score (\APS):}
{\APS} is computed as
the average property score for each molecular property across 
all successfully optimized molecules.
Higher or lower {\APS}, depending on the desired direction for each property, indicates that the model consistently generates
better molecules with property scores aligned with pharmaceutical objectives.

\end{enumerate}

\section{Prompt Templates}
\label{sec:app:prompt}

The prompt templates for general-purpose LLMs and for \LlaSMol
are provided below.

\subsection{Prompt Template for General-purpose LLMs}
\label{sec:app:prompt:glm}

We use a structured and detailed prompt template with a system prompt, task instruction, and in-context examples for few-shot prompting.
Figure~\ref{fig:prompt_glm} shows an example.

\begin{figure*}[!t]
\begin{tcolorbox}[
  colback=lightergray, 
  colframe=black, 
  sharp corners, 
  boxrule=0.5pt, 
  width=\textwidth,
  left=1mm, 
  right=1mm, 
  top=1mm, 
  bottom=1mm 
]
\begin{lstlisting}[
  language=, 
  basicstyle=\ttfamily\footnotesize, 
  breaklines=true, 
  breakindent=0pt, % Disable indentation for wrapped lines
  showstringspaces=false, % Remove visible spaces
  xleftmargin=0pt, % Remove left margin
  xrightmargin=0pt, % Remove right margin
  aboveskip=0pt,belowskip=0pt
]
<<SYS>>
You are an expert medicinal chemist specializing in molecular optimization. You understand how structural modifications affect key ADMET properties and inhibitions of common receptor targets like DRD2.
<</SYS>>

[INST]
Your task is to modify the given molecule to adjust specific molecular properties while keeping structural changes as minimal as possible. Use the examples (if provided) as a guide. Your response should only contain a valid  SMILES representation of the modified molecule enclosed with <SMILES> </SMILES> tag.

Examples:
%%% Input : <SMILES> O=C(Cc1cccc([N+](=O)[O-])c1)NC1CCN(Cc2ccccc2)CC1 </SMILES>
%%% Adjust: increase DRD2 inhibition with a goal of at least <THRESHOLD> 0.54 </THRESHOLD>, decrease Mutagenicity with a goal of at most <THRESHOLD> 0.1 </THRESHOLD> and increase QED aiming for at least <THRESHOLD> 0.89 </THRESHOLD> while keeping Intestinal adsorption unchanged.
%%% Response: <SMILES> O=C(Cc1ccc(O)cc1)NC1CCN(Cc2ccccc2)CC1 </SMILES>

Task:
%%% Input : <SMILES> C#Cc1ccc(C2CC3CCC(C2C(=O)OC)N3C)cc1 </SMILES>
%%% Adjust: decrease Mutagenicity with a goal of at most <THRESHOLD> 0.2 </THRESHOLD>, increase QED with a goal of at least <THRESHOLD> 0.8 </THRESHOLD> and increase the value of DRD2 inhibition to be at least <THRESHOLD> 0.2 </THRESHOLD> while keeping Intestinal adsorption unchanged.
[/INST]

%%% Response:
\end{lstlisting}
\end{tcolorbox}
\vspace{-10pt}
\caption{An example of a prompt used for general-purpose LLMs}
\label{fig:prompt_glm}
\end{figure*}

\subsection{Prompt Template for \LlaSMol}
\label{sec:app:prompt:llasmol}

Unlike general-purpose language models, \LlaSMol was instruction-tuned on a range of chemistry-specific tasks using a dedicated prompt structure. 
In our preliminary experiments, 
we found that applying the general-purpose prompt format led to suboptimal performance, as \LlaSMol often failed to interpret the task correctly. 
To address this, we adopted a simplified prompt format that omits the system message and does not explicitly separate the instruction, input, and expected output. 
Additionally, we restrict our evaluation of \LlaSMol to 0-shot inference only.
Figure~\ref{fig:prompt_llasmol} illustrates the simplified prompt used for the same task as above.

\begin{figure*}
\begin{tcolorbox}[
  colback=lightergray, 
  colframe=black, 
  sharp corners, 
  boxrule=0.5pt, 
  width=\textwidth,
  left=1mm, 
  right=1mm, 
  top=1mm, 
  bottom=1mm 
]
\begin{lstlisting}[
  language=, 
  basicstyle=\ttfamily\footnotesize, 
  breaklines=true, 
  breakindent=0pt, % Disable indentation for wrapped lines
  showstringspaces=false, % Remove visible spaces
  xleftmargin=0pt, % Remove left margin
  xrightmargin=0pt, % Remove right margin
  aboveskip=0pt,belowskip=0pt
]
Modify the molecule <SMILES> C#Cc1ccc(C2CC3CCC(C2C(=O)OC)N3C)cc1 <SMILES> to decrease the value of Mutagenicity to be at most <THRESHOLD> 0.2 </THRESHOLD>, increase QED to be at least <THRESHOLD> 0.8 </THRESHOLD> and increase DRD2 inhibition to be at least <THRESHOLD> 0.2 </THRESHOLD> while keeping Intestinal adsorption unchanged.
%%% Response:
\end{lstlisting}
\end{tcolorbox}
\vspace{-10pt}
\caption{An example of a prompt used for \LlaSMol}
\label{fig:prompt_llasmol}
\end{figure*}

\begin{table*}[h]
\footnotesize
\centering
\caption{Licenses and Sources of Artifacts}
\label{tbl:baseline_LLMs}
\setlength{\tabcolsep}{0pt}%

\begin{small}
\begin{threeparttable}

\begin{tabular}{
    @{\hspace{2pt}}l@{\hspace{2pt}}
    @{\hspace{2pt}}p{0.45\textwidth}@{\hspace{2pt}}
    @{\hspace{2pt}}p{0.2\textwidth}@{\hspace{2pt}}
    @{\hspace{2pt}}l@{\hspace{2pt}}
}

\toprule
\textbf{Artifact} & \textbf{Source} & \textbf{License Type} & \textbf{Accessibility} \\
\midrule

\multirow{2}{*}{Modof} & \url{https://github.com/ziqi92/Modof} & {PolyForm Noncommercial License 1.0.0} & Open Source\\

\multirow{2}{*}{\LlaSMolM} & \url{https://huggingface.co/datasets/osunlp/SMolInstruct} & {Creative Commons Attribution 4.0} & \multirow{2}{*}{Checkpoint}\\

\multirow{2}{*}{\ChemDFML} &
\url{https://huggingface.co/OpenDFM/ChemDFM-v1.5-8B} &
GNU Affero General Public License v3.0 
& \multirow{2}{*}{Checkpoint}\\

\multirow{2}{*}{Claude 3.5 (Sonnet)} & \url{https://docs.anthropic.com/claude/reference/getting-started-with-the-api} & 
\multirow{2}{*}{Proprietary} 
& \multirow{2}{*}{API} \\

GPT-4o & \url{https://openai.com/api/} & 
Proprietary
& API \\

\multirow{2}{*}{Llama-3.1 8B-Instruct} & \url{https://huggingface.co/meta-llama/Llama-3.1-8B-Instruct} & \multirow{2}{*}{Llama 3.1 Community} & \multirow{2}{*}{Checkpoint} \\

\multirow{2}{*}{Mistral-7B-Instruct-v0.3} & \url{https://huggingface.co/mistralai/Mistral-7B-Instruct-v0.3} & \multirow{2}{*}{Apache license 2.0} & \multirow{2}{*}{Checkpoint} \\
\bottomrule
\end{tabular}

\end{threeparttable}
\end{small}

\end{table*}

\section{Case Studies}
\label{sec:app:case_study}

\subsection{Case from \ACEP}
\label{sec:app:case_study:acep}
\begin{figure}[h!]
    \centering
    \begin{subfigure}{0.48\textwidth}
    \includegraphics[width=\textwidth]{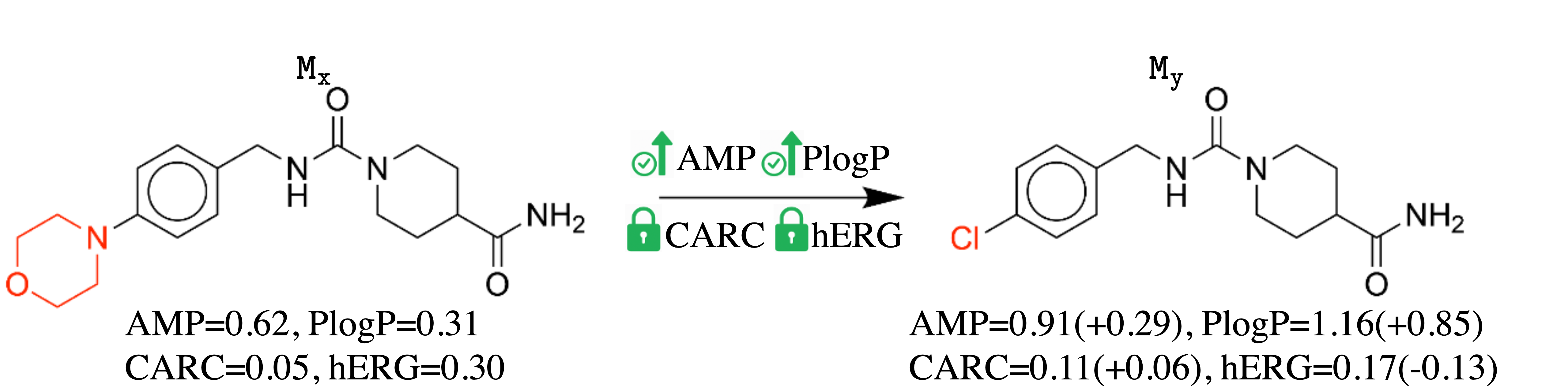}
    \caption{\mollmDecGenM optimization}
    \label{fig:ACEP_case_method}
    \end{subfigure}
    \hfill
    \begin{subfigure}{0.48\textwidth}
    \centering
    \includegraphics[width=\textwidth]{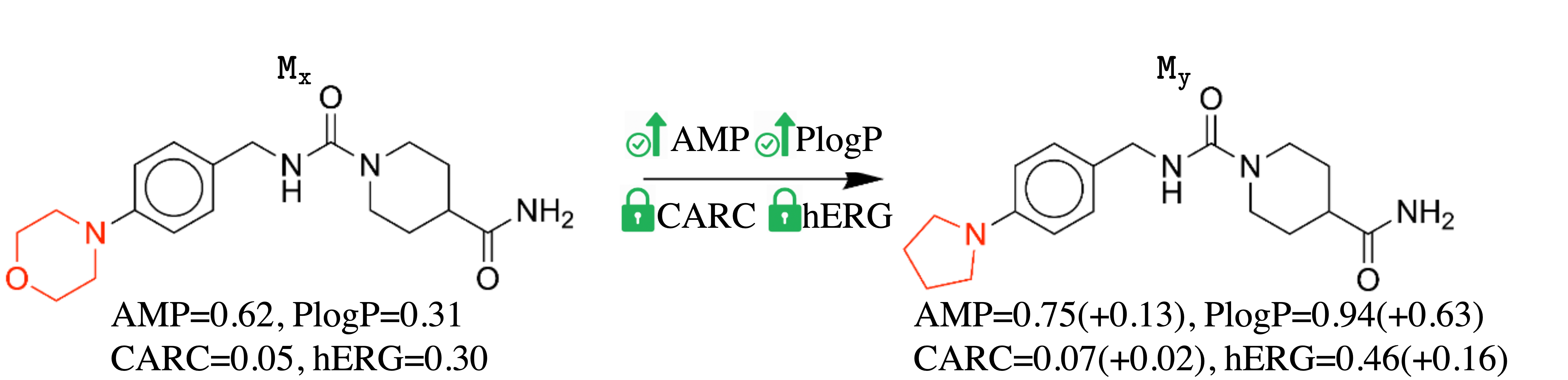}
    \caption{\LlaSMolM optimization}
    \label{fig:ACEP_case_llasmol}
    \end{subfigure}
    \caption{An example from \ACEP. Modifications are highlighted in red.}
\end{figure}

Figure~\ref{fig:ACEP_case_method} and Figure~\ref{fig:ACEP_case_llasmol} show optimization examples generated by \mollmDecGenM and \LlaSMolM on the IND task \ACEP.
The hit molecule features a central urea scaffold with a carboxamide and a morpholine ring. The goal is to improve AMP and PlogP while maintaining CARC and hERG.
%

\mollmDecGenM accomplishes this by replacing the morpholine with a para-chlorophenyl group (Figure~\ref{fig:ACEP_case_method}).
This modification eliminates a polar heterocycle and introduces a planar, lipophilic aromatic ring bearing a chlorine atom. 
This leads to notable improvements in AMP (+0.29) 
and PlogP (+0.85), while CARC and hERG remain within acceptable ranges.
The increased hydrophobicity introduced by the chlorinated aromatic ring contributes to a higher PlogP, 
as aromatic chlorides are known to enhance lipophilicity 
due to both the non-polar nature of the phenyl group and the electron-withdrawing effect of chlorine~\cite{hansch1995exploring}.
The rigid aromatic system may reduce the molecule’s conformational flexibility, which in turn lowers conformational entropy. 
This structural constraint can limit the number of unintended binding interactions, thereby reducing the likelihood of off-target liabilities~\cite{meanwell2011synopsis,meanwell2016improving}

\LlaSMolM's modification replaces the morpholine with a pyrrolidine ring. This change maintains a basic nitrogen atom but removes the oxygen, slightly reducing polarity compared to morpholine. 
Although this approach achieves a moderate PlogP improvement (+0.63), it shows a concerning increase in hERG liability (+0.16). 
The pyrrolidine ring, while structurally similar to morpholine (Figure~\ref{fig:ACEP_case_llasmol}), introduces greater basicity and conformational flexibility. 
These properties are known risk factors for hERG channel binding in medicinal chemistry, explaining the less favorable safety profile~\cite{cavalli2002toward}.

\begin{figure}[h!]
\begin{subfigure}{0.48\textwidth}
    \centering
    \includegraphics[width=\textwidth]{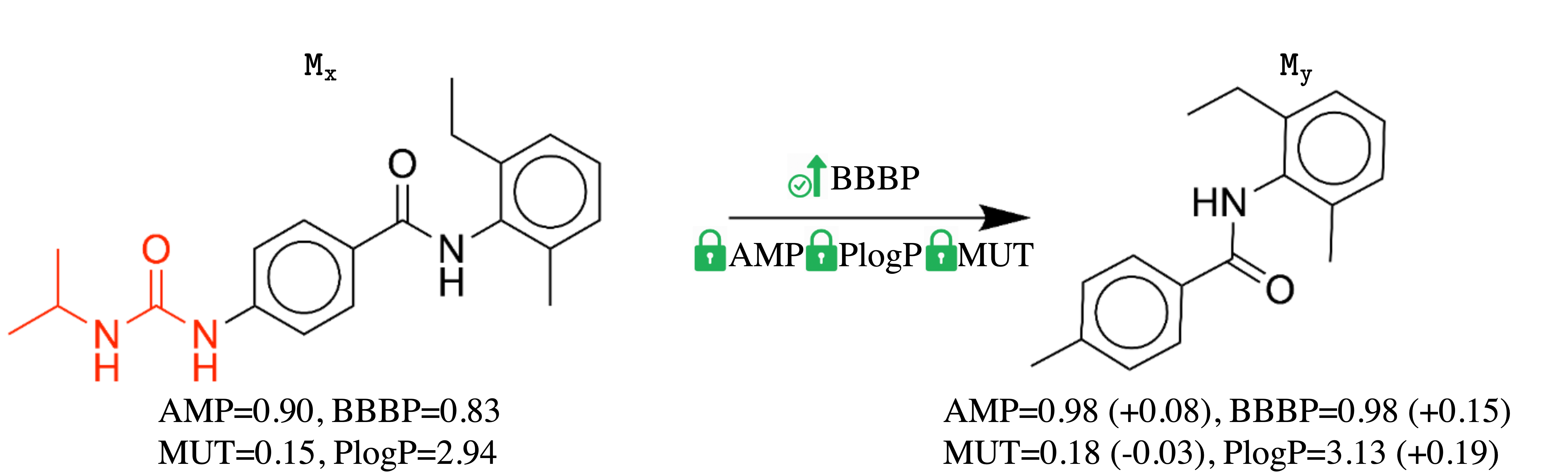}
    \caption{\mollmDecGenM optimization}
    \label{fig:ABMP_case_method}
\end{subfigure}

\begin{subfigure}{0.48\textwidth}
    \centering
    \includegraphics[width=\textwidth]{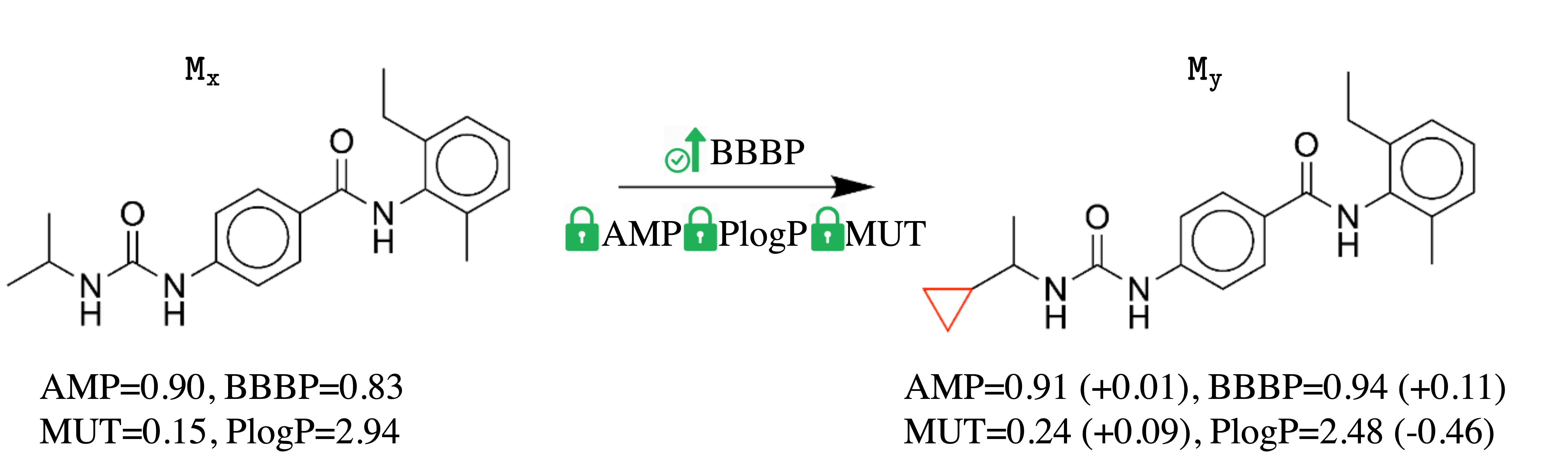}
    \caption{\LlaSMolM optimization}
    \label{fig:ABMP_case_llasmol}
\end{subfigure}
\caption{An example from \ABMP. Modifications are highlighted in red. }
\end{figure}
\subsection{Case from \ABMP}
\label{sec:app:case_study:abmp}
Figure~\ref{fig:ABMP_case_method} and Figure~\ref{fig:ABMP_case_llasmol} present optimization examples produced by \mollmDecGenM and \LlaSMolM on the OOD task \ABMP.
The hit molecule is a symmetric tri-amide structure, composed of three carbonyl linkers connecting aromatic and aliphatic moieties. The goal is to improve BBBP, while keeping AMP, MUT, and PlogP stable.

\mollmDecGenM introduces a substantial simplification by collapsing the tri-amide backbone into a more compact structure containing a single central amide and two substituted aromatic rings (Figure~\ref{fig:ABMP_case_method}). This transformation removes several polar functional groups and incorporates lipophilic features such as methyl and aryl substitutions. These changes are well-aligned with medicinal chemistry strategies for enhancing membrane permeability -- primarily through increased lipophilicity and reduced polarity~\cite{meanwell2011improving,leeson2007influence}. 
As a result, \mollmDecGenM achieves a favorable outcome, yielding a significant improvement in BBBP (+0.15), along with a modest increase in PlogP (+0.19), while keeping AMP and MUT values stable.

In contrast, \LlaSMolM applies a conservative modification by retaining the tri-amide scaffold and appending an isopropyl group to the left-hand side of the molecule (Figure~\ref{fig:ABMP_case_llasmol}). 
This change preserves the molecule’s original polarity and structural complexity, while introducing additional steric bulk. Crucially, it fails to reduce polarity or increase hydrophobicity
-- both essential for maintaining or improving PlogP~\cite{ertl2009estimation}. 
As a result, despite a small gain in BBBP (+0.11), the model suffers a substantial drop in PlogP (–0.46) and an increase in toxicity (MUT), indicating an unfavorable optimization outcome.

\section{Complete Experimental Results}
\label{sec:app:results}

\subsection{IND Evaluation}
\label{sec:app:results:ind}

Tables~\ref{tbl:bpq_ind},~\ref{tbl:elq_ind},~\ref{tbl:acep_ind},~\ref{tbl:bdpq_ind} and \ref{tbl:dhmq_ind} presents the performance comparison
of {\mollm}s with general-purpose LLMs and \LlaSMolM under all evaluation metrics for each IND task.

Table~\ref{tbl:main_ind_strict} presents the overall performance
comparison of {\mollm}s with all baselines under the strict success criteria.
This requires each sub-optimal property to exceed its predefined pharmaceutically relevant threshold, $\Theta_p$, in the optimized molecule.
We use $\Theta_p$ to reflect realistic drug design objectives,
where each property is expected to reach a clinically meaningful level. 
However, this is a highly challenging setting, particularly because our evaluation involves only a single-step molecule modification. 
Starting molecules may be significantly sub-optimal, and a single structural change may not be sufficient to reach such high thresholds.
This explains the significantly lower success rates for all models compared
to the looser success criteria in Table~\ref{tbl:main_ind}.

\begin{table*}[h!]
\centering
\setlength{\tabcolsep}{0pt}%
\caption{Overall Performance in IND Tasks with stricter success criteria}
\label{tbl:main_ind_strict}
\vspace{-5pt}
\begin{small}
\begin{threeparttable}
\begin{tabular}{
   @{\hspace{0pt}}l@{\hspace{5pt}}
   @{\hspace{2pt}}r@{\hspace{2pt}}
   @{\hspace{2pt}}r@{\hspace{2pt}}
   @{\hspace{2pt}}r@{\hspace{2pt}}
   @{\hspace{5pt}}c@{\hspace{5pt}} 
   @{\hspace{0pt}}r@{\hspace{2pt}}
   @{\hspace{2pt}}r@{\hspace{2pt}}
   @{\hspace{2pt}}r@{\hspace{2pt}}
   @{\hspace{5pt}}c@{\hspace{5pt}} 
   @{\hspace{0pt}}r@{\hspace{2pt}}
   @{\hspace{2pt}}r@{\hspace{2pt}}
   @{\hspace{2pt}}r@{\hspace{2pt}}
   @{\hspace{5pt}}c@{\hspace{5pt}} 
   @{\hspace{0pt}}r@{\hspace{2pt}}
   @{\hspace{2pt}}r@{\hspace{2pt}}
   @{\hspace{2pt}}r@{\hspace{2pt}}
   @{\hspace{5pt}}c@{\hspace{5pt}} 
   @{\hspace{0pt}}r@{\hspace{2pt}}
   @{\hspace{2pt}}r@{\hspace{2pt}}
   @{\hspace{2pt}}r@{\hspace{0pt}}
}
\toprule
\multirow{2}{*}{Model} & \multicolumn{3}{c}{\BPQ} && \multicolumn{3}{c}{\ELQ} && \multicolumn{3}{c}{\ACEP}
&& \multicolumn{3}{c}{\BDPQ}  && \multicolumn{3}{c}{\DHMQ} \\
\cmidrule(){2-4} \cmidrule(){6-8} \cmidrule(){10-12} \cmidrule(){14-16} \cmidrule(){18-20}
& \SSR$^{\uparrow}$ & \Sim$^{\uparrow}$ & \RI$^{\uparrow}$ & 
& \SSR$^{\uparrow}$ & \Sim$^{\uparrow}$ & \RI$^{\uparrow}$ &
& \SSR$^{\uparrow}$ & \Sim$^{\uparrow}$ & \RI$^{\uparrow}$ &
& \SSR$^{\uparrow}$ & \Sim$^{\uparrow}$ & \RI$^{\uparrow}$ &
& \SSR$^{\uparrow}$ & \Sim$^{\uparrow}$ & \RI$^{\uparrow}$ \\
\midrule

\rowcolor{lightgray}
\multicolumn{20}{c}{\textbf{General-purpose LLMs}} 
\\
Mistral (0-shot) & 3.40 & 0.71 & \underline{1.60} &  & 3.40 & \textbf{\underline{0.70}} & 0.38 &  & 2.80 & 0.70 & 0.88 &  & 0.00 &- & - &  & 0.00 &-  &-  \\
Llama (0-shot) & 3.80 & 0.69 & 0.39 &  & 2.20 & 0.69 & 0.27 &  & 1.00 & 0.71 & 0.53 &  & 0.00 & - & - &  & 0.20 & {0.75} & 3.00 \\
Claude-3.5 (0-shot) & 4.40 & 0.65 & 0.56 &  & 3.00 & 0.63 & 0.40 &  & 1.60 & 0.60 & 0.72 &  & 0.00 & - & - &  & 0.00 & - &-  \\
GPT-4o (0-shot) & 1.60 & \textbf{\underline{0.73}} & 0.48 &  & 1.40 & 0.67 & 0.33 &  & 1.60 & 0.72 & 0.34 &  & 0.00 &- &-  &  & 0.40 & 0.71 & 2.51 \\
\cellcolor{yellow!20}Mistral (1-shot) & 14.20 & 0.53 & 1.45 &  & 16.20 & 0.57 & \underline{0.49} &  & 10.20 & 0.54 & \underline{1.31} &  & \underline{\cellcolor{yellow!20}3.40} & \cellcolor{yellow!20}0.32 & \cellcolor{yellow!20}18.68 &  & \underline{\cellcolor{yellow!20}3.40} & \cellcolor{yellow!20}0.39 & \cellcolor{yellow!20}3.87 \\
Llama (1-shot) & 6.40 & 0.63 & 0.62 &  & 4.80 & 0.61 & 0.39 &  & 3.00 & 0.63 & 0.47 &  & 0.40 & 0.15 & \underline{18.71} &  & 2.20 & 0.28 & \textbf{\underline{14.00}} \\
Claude-3.5 (1-shot) & 9.20 & 0.59 & 0.95 &  & 3.20 & 0.63 & 0.42 &  & 3.60 & \textbf{\underline{0.73}} & 0.72 &  & 0.60 & 0.38 & 4.16 &  & 0.40 & 0.69 & 2.73 \\
GPT-4o (1-shot) & 2.60 & 0.70 & 0.45 &  & 2.00 & 0.67 & 0.28 &  & 1.20 & \textbf{\underline{0.73}} & 0.25 &  & 0.00 &- & - &  & 1.00 & 0.71 & 2.72 \\

\rowcolor{lightgray}
\multicolumn{20}{c}{\textbf{Foundational LLMs for Chemistry}}
\\
\cellcolor{yellow!20}LlaSMol-M & \underline{\cellcolor{yellow!20}14.80} & \cellcolor{yellow!20}0.61 & \cellcolor{yellow!20}0.88 &  & \underline{\cellcolor{yellow!20}17.60} & \cellcolor{yellow!20}0.60 & \cellcolor{yellow!20}0.48 &  & \underline{\cellcolor{yellow!20}10.80} & \cellcolor{yellow!20}0.62 & \cellcolor{yellow!20}0.67 &  & 0.60 & \textbf{\underline{0.68}} & 9.42 &  & 1.40 & 0.70 & 4.12 \\

\ChemDFML & 3.20 & 0.63 & 0.33 &  & 3.00 & 0.65 & 0.38 &  & 1.40 & 0.69 & 0.40 &  & 0.20 & 0.55 & 0.78 &  & 0.60 & \textbf{\underline{0.81}} & 5.44 \\

\rowcolor{lightgray}
\multicolumn{20}{c}{\textbf{Specialist LLMs}}
\\

\cellcolor{green!10}\mollmTaskM & 25.40 & 0.51 & 2.57 &  
& 28.80 & 0.51 & 0.56 &  
& 28.00 & 0.50 & \textbf{4.00} &  
& \textbf{\cellcolor{green!10}9.40} & \cellcolor{green!10}0.35 & \cellcolor{green!10}13.24 &  
& \cellcolor{green!10}6.40 & \cellcolor{green!10}0.52 & \cellcolor{green!10}9.92 \\

\cellcolor{green!10}\mollmTaskL 
& \cellcolor{green!10}29.60 & \cellcolor{green!10}0.53 & \cellcolor{green!10}2.06 & 
& \cellcolor{green!10}31.40 & \cellcolor{green!10}0.50 & \textbf{\cellcolor{green!10}0.58} &  
& \cellcolor{green!10}31.40 & \cellcolor{green!10}0.50 & \cellcolor{green!10}3.14 &  
& 4.60 & 0.48 & 16.89 &  
& 4.20 & 0.65 & 10.68 \\

\hline

\ImpT (\%) & 100.0 & -13.1 & 134.1 &  & 78.4 & -16.7 & 20.8 &  & 190.7 & -19.4 & 368.7 &  & 176.5 & 9.4 & -29.1 &  & 88.2 & 33.3 & 156.3 \\

\rowcolor{lightgray}
\multicolumn{20}{c}{\textbf{Generalist LLMs}}
\\

\cellcolor{blue!10}\mollmNGenM
& 27.60 & 0.59 & 2.43 &  
& 23.40 & 0.62 & 0.51 &  
& 31.20 & 0.57 & 3.42 &  
& 5.40 & 0.55 & 11.30 &  
& \textbf{\cellcolor{blue!10}9.00} & \cellcolor{blue!10}0.54 & \cellcolor{blue!10}11.53 \\

\cellcolor{blue!10}\mollmNGenL 
& 30.60 & 0.57 & 2.15 &  
& 25.60 & 0.60 & 0.51 &  
& \textbf{\cellcolor{blue!10}34.40} & \cellcolor{blue!10}0.55 & \cellcolor{blue!10}2.77 &  
& \cellcolor{blue!10}6.40 & \cellcolor{blue!10}0.50 & \cellcolor{blue!10}19.46 &  
& 6.80 & 0.60 & 13.35 \\

\cellcolor{blue!10}\mollmDecGenM & \textbf{\cellcolor{blue!10}32.60} & \cellcolor{blue!10}0.59 & \cellcolor{blue!10}2.32 &  & \textbf{\cellcolor{blue!10}32.00} & \cellcolor{blue!10}0.57 & \cellcolor{blue!10}0.55 &  & 23.40 & 0.58 & 1.88 &  & 3.80 & 0.59 & 13.26 &  & 4.80 & 0.64 & 11.14 \\

\mollmDecGenL & 32.40 & 0.54 & \textbf{2.59} &  & 27.60 & 0.56 & 0.54 &  & 25.20 & 0.56 & 3.11 &  & 5.00 & 0.51 & \textbf{22.70} &  & 5.40 & 0.56 & 13.70 \\
\hline
\ImpG (\%) & 120.3 & -3.3 & 163.6 &  & 81.8 & -5.0 & 14.6 &  & 218.5 & -11.3 & 313.4 &  & 88.2 & 56.2 & 4.2 &  & 164.7 & 38.5 & 197.9 \\

\bottomrule
\end{tabular}

\begin{tablenotes}[normal,flushleft]
\footnotesize
\item $^\uparrow$ and $^\downarrow$ indicate whether a higher or lower value of the metric is preferred, respectively.
For each task, we \underline{underline} the best baseline performance
and highlight in \textbf{bold} the best performing model for each metric.
{\ImpT} and {\ImpG} represent the relative percentage improvement from the \colorbox{green!10}{best specialist LLM} 
and \colorbox{blue!10}{best generalist LLM} over the 
\colorbox{yellow!20}{best baseline},
respectively.
The best model in each group is selected based on {\SR} for each task.
\end{tablenotes}

\end{threeparttable}
\end{small}
\end{table*}

\begin{table*}[h!]
\centering
\caption{Overall Performance on \BPQ}
\setlength{\tabcolsep}{0pt}%
\label{tbl:bpq_ind}
\begin{small}
\begin{threeparttable}

\begin{tabular}{
    @{\hspace{9pt}}l@{\hspace{9pt}}
    @{\hspace{9pt}}r@{\hspace{9pt}}
    @{\hspace{9pt}}r@{\hspace{9pt}}
    @{\hspace{9pt}}r@{\hspace{9pt}}
    @{\hspace{9pt}}r@{\hspace{9pt}}
    @{\hspace{9pt}}r@{\hspace{9pt}}
    @{\hspace{9pt}}r@{\hspace{9pt}}
    @{\hspace{4pt}}r@{\hspace{4pt}}
    @{\hspace{4pt}}r@{\hspace{4pt}}
    @{\hspace{4pt}}r@{\hspace{4pt}}
}
\toprule
\multirow{2}{*}{Model} 
& \multirow{2}{*}{\SR$^{\uparrow}$}
& \multirow{2}{*}{\Val$^{\uparrow}$} 
& \multirow{2}{*}{\Sim$^{\uparrow}$} 
& \multirow{2}{*}{\Nov$^{\uparrow}$}
& \multirow{2}{*}{\SAS$^{\downarrow}$}
& \multirow{2}{*}{\RI$^{\uparrow}$}
& \multicolumn{3}{c}{\APS}
\\
\cmidrule(){8-10} 
& & & & & & & BBBP$^\uparrow$ & PlogP$^\uparrow$ & QED$^\uparrow$ 
\\
\midrule

\rowcolor{lightgray}
\multicolumn{10}{c}{\textbf{General-purpose LLMs}} 
\\
Mistral (0-shot) & 28.80 & 85.80 & \textbf{\underline{0.75}} & \textbf{\underline{100.00}} & 2.87 & 1.24 & 0.92 & 0.41 & \underline{0.77} \\
Llama (0-shot) & 33.60 & 99.00 & 0.70 & \textbf{\underline{100.00}} & 2.86 & 0.78 & 0.92 & 0.65 & 0.76 \\
Claude-3.5 (0-shot) & 51.80 & 96.80 & 0.68 & 99.61 & 2.75 & 0.89 & 0.91 & 0.70 & 0.75 \\
GPT-4o (0-shot) & 30.20 & 88.00 & 0.72 & \textbf{\underline{100.00}} & 2.70 & 0.55 & 0.90 & 0.65 & 0.76 \\
Mistral (1-shot) & 72.80 & 99.20 & 0.63 & 97.53 & \underline{2.58} & 1.26 & 0.91 & \underline{1.07} & \underline{0.77} \\
Llama (1-shot) & 49.60 & \textbf{\underline{100.00}} & 0.68 & 99.19 & 2.71 & 0.95 & 0.91 & 0.89 & 0.75 \\
Claude-3.5 (1-shot) & 61.80 & 96.60 & 0.65 & \textbf{\underline{100.00}} & 2.68 & \underline{1.31} & \textbf{\underline{0.93}} & 0.90 & \underline{0.77} \\
GPT-4o (1-shot) & 28.60 & 86.20 & 0.74 & \textbf{\underline{100.00}} & 2.76 & 0.77 & 0.90 & 0.70 & 0.76 \\

\rowcolor{lightgray}
\multicolumn{10}{c}{\textbf{Foundational LLMs for Chemistry}} 
\\

\cellcolor{yellow!20}\LlaSMolM & \underline{\cellcolor{yellow!20}78.20} & \textbf{\underline{\cellcolor{yellow!20}100.00}} & \cellcolor{yellow!20}0.64 & \cellcolor{yellow!20}99.74 & \cellcolor{yellow!20}2.65 & \cellcolor{yellow!20}0.92 & \cellcolor{yellow!20}0.91 & \cellcolor{yellow!20}0.87 & \underline{\cellcolor{yellow!20}0.77} \\

\ChemDFML & 27.00 & 92.00 & 0.66 & 99.26 & 2.82 & 0.65 & \textbf{\underline{0.93}} & 0.68 & \underline{0.77} \\

\rowcolor{lightgray}
\multicolumn{10}{c}{\textbf{Specialist LLMs}}
\\
\mollmTripleTaskM & 71.00 & 98.40 & 0.57 & 98.87 & 2.45 & 2.59 & \textbf{0.93} & 1.51 & \textbf{0.79} \\
\cellcolor{green!10}\mollmTripleTaskL & \cellcolor{green!10}84.20 & \textbf{\cellcolor{green!10}100.00} & \cellcolor{green!10}0.58 & \cellcolor{green!10}99.05 & \cellcolor{green!10}2.46 & \cellcolor{green!10}2.09 & \cellcolor{green!10}0.92 & \cellcolor{green!10}1.44 & \textbf{\cellcolor{green!10}0.79} \\


\hline
\ImpT & 7.7 & 0.0 & -9.4 & -0.7 & 7.2 & 127.2 & 1.1 & 65.5 & 2.6 \\

\rowcolor{lightgray}
\multicolumn{10}{c}{\textbf{Generalist LLMs}} 
\\

\mollmTripleGenM & 84.80 & \textbf{100.00} & 0.63 & 99.06 & 2.46 & 2.64 & 0.92 & 1.47 & 0.78 \\
\mollmTripleGenL  & 88.80 & \textbf{100.00} & 0.62 & 99.10 & \textbf{2.38} & 2.16 & 0.92 & 1.48 & \textbf{0.79} \\



\cellcolor{blue!10}\mollmDecGenM & \textbf{\cellcolor{blue!10}89.40} & \cellcolor{blue!10}99.00 & \cellcolor{blue!10}0.62 & \cellcolor{blue!10}98.43 & \cellcolor{blue!10}2.49 & \cellcolor{blue!10}2.30 & \textbf{\cellcolor{blue!10}0.93} & \cellcolor{blue!10}1.39 & \textbf{\cellcolor{blue!10}0.79} \\

\mollmDecGenL & 79.40 & 88.80 & 0.57 & 97.48 & 2.42 & \textbf{2.67} & \textbf{0.93} & \textbf{1.56} & \textbf{0.79} \\

\hline
\ImpG & 14.3 & -1.0 & -3.1 & -1.3 & 6.0 & 150.0 & 2.2 & 59.8 & 2.6 \\

\bottomrule
\end{tabular}

\begin{tablenotes}[normal,flushleft]
\footnotesize
\item $^\uparrow$ and $^\downarrow$ indicate whether a higher or lower value of the metric is preferred, respectively.
For each task, we \underline{underline} the best baseline performance
and highlight in \textbf{bold} the best performing model for each metric.
{\ImpT} and {\ImpG} represent the relative percentage improvement from the \colorbox{green!10}{best specialist LLM} 
and \colorbox{blue!10}{best generalist LLM} over the 
\colorbox{yellow!20}{best baseline},
respectively.
The best model in each group is selected based on {\SR} for each task.
\end{tablenotes}

\end{threeparttable}
\end{small}
\end{table*}
\begin{table*}[h!]
\centering
\caption{Overall Performance on \ELQ}
\setlength{\tabcolsep}{0pt}%
\label{tbl:elq_ind}
\begin{small}
\begin{threeparttable}

\begin{tabular}{
    @{\hspace{9pt}}l@{\hspace{9pt}}
    @{\hspace{9pt}}r@{\hspace{9pt}}
    @{\hspace{9pt}}r@{\hspace{9pt}}
    @{\hspace{9pt}}r@{\hspace{9pt}}
    @{\hspace{9pt}}r@{\hspace{9pt}}
    @{\hspace{9pt}}r@{\hspace{9pt}}
    @{\hspace{9pt}}r@{\hspace{9pt}}
    @{\hspace{4pt}}r@{\hspace{4pt}}
    @{\hspace{4pt}}r@{\hspace{4pt}}
    @{\hspace{4pt}}r@{\hspace{4pt}}
}
\toprule
\multirow{2}{*}{Model} 
& \multirow{2}{*}{\SR$^{\uparrow}$}
& \multirow{2}{*}{\Val$^{\uparrow}$} 
& \multirow{2}{*}{\Sim$^{\uparrow}$} 
& \multirow{2}{*}{\Nov$^{\uparrow}$}
& \multirow{2}{*}{\SAS$^{\downarrow}$}
& \multirow{2}{*}{\RI$^{\uparrow}$}
& \multicolumn{3}{c}{\APS}
\\
\cmidrule(){8-10} 
& & & & & & & hERG$^\downarrow$ & LIV$^\downarrow$ & QED$^\uparrow$ 
\\
\midrule

\rowcolor{lightgray}
\multicolumn{10}{c}{\textbf{General-purpose LLMs}} 
\\
Mistral (0-shot) & 21.60 & 89.20 & 0.72 & \textbf{\underline{100.00}} & 2.82 & 0.16 & \underline{0.37} & 0.55 & 0.77 \\
Llama (0-shot) & 16.60 & 97.40 & \textbf{\underline{0.74}} & \textbf{\underline{100.00}} & 2.90 & 0.10 & 0.44 & 0.56 & \underline{0.80} \\
Claude-3.5 (0-shot) & 20.00 & 96.40 & 0.64 & \textbf{\underline{100.00}} & \underline{2.67} & 0.20 & 0.41 & 0.60 & 0.76 \\
GPT-4o (0-shot) & 16.60 & 90.80 & 0.72 & \textbf{\underline{100.00}} & 2.83 & 0.10 & 0.39 & 0.53 & 0.74 \\
Mistral (1-shot) & 74.80 & \underline{99.80} & 0.59 & 94.92 & 2.77 & \underline{0.28} & 0.38 & 0.55 & 0.78 \\
Llama (1-shot) & 36.80 & 99.40 & 0.68 & 97.83 & 2.90 & 0.15 & 0.45 & 0.56 & 0.77 \\
Claude-3.5 (1-shot) & 29.20 & 97.60 & 0.63 & \textbf{\underline{100.00}} & 2.73 & 0.21 & 0.48 & 0.58 & 0.76 \\
GPT-4o (1-shot) & 19.60 & 90.00 & 0.72 & \textbf{\underline{100.00}} & 2.85 & 0.12 & 0.46 & 0.53 & 0.76 \\

\rowcolor{lightgray}
\multicolumn{10}{c}{\textbf{Foundational LLMs for Chemistry}} 
\\

\cellcolor{yellow!20}\LlaSMolM   & \underline{\cellcolor{yellow!20}81.40} & \underline{\cellcolor{yellow!20}99.80} & \cellcolor{yellow!20}0.62 & \cellcolor{yellow!20}99.26 & \cellcolor{yellow!20}2.71 & \underline{\cellcolor{yellow!20}0.28} & \cellcolor{yellow!20}0.38 & \cellcolor{yellow!20}0.56 & \cellcolor{yellow!20}0.77 \\

\ChemDFML & 15.00 & 91.20 & 0.68 & \textbf{\underline{100.00}} & 2.91 & 0.19 & 0.38 & \underline{0.52} & 0.79 \\

\rowcolor{lightgray}
\multicolumn{10}{c}{\textbf{Specialist LLMs}}
\\
\mollmTripleTaskM  
& 81.80 & 99.40 & 0.55 & 99.27 & 2.85 & 0.39 & 0.32 & \textbf{0.46} & 0.79 \\

\cellcolor{green!10}\mollmTripleTaskL 
& \cellcolor{green!10}85.40 & \textbf{\cellcolor{green!10}100.00} & \cellcolor{green!10}0.53 & \cellcolor{green!10}99.53 & \cellcolor{green!10}2.87 & \textbf{\cellcolor{green!10}0.41} & \textbf{\cellcolor{green!10}0.29} & \textbf{\cellcolor{green!10}0.46} & \cellcolor{green!10}0.79 \\


\hline
\ImpT 
& 4.9 & 0.2 & -14.5 & 0.3 & -5.9 & 46.4 & 23.7 & 17.9 & 2.6 \\

\rowcolor{lightgray}
\multicolumn{10}{c}{\textbf{Generalist LLMs}} 
\\

\mollmTripleGenM & 83.20 & 99.80 & 0.63 & 98.80 & 2.64 & 0.33 & 0.33 & 0.53 & 0.78 

\\
\cellcolor{blue!10}\mollmTripleGenL 
& \textbf{\cellcolor{blue!10}90.80} & \textbf{\cellcolor{blue!10}100.00} & \cellcolor{blue!10}0.63 & \cellcolor{blue!10}98.90 & \cellcolor{blue!10}2.60 & \cellcolor{blue!10}0.34 & \cellcolor{blue!10}0.33 & \cellcolor{blue!10}0.52 & \cellcolor{blue!10}0.80 \\

\mollmDecGenM 
& 88.40 & 99.80 & 0.59 & 99.55 & 2.64 & \textbf{0.41} & \textbf{0.29} & 0.50 & \textbf{0.81} \\

\mollmDecGenL 
 & 79.00 & 90.60 & 0.56 & 99.49 & \textbf{2.58} & \textbf{0.41} & 0.30 & 0.48 & \textbf{0.81} \\

\hline
\ImpG 
& 11.5 & 0.2 & 1.6 & -0.4 & 4.1 & 21.4 & 13.2 & 7.1 & 3.9 \\

\bottomrule
\end{tabular}

\begin{tablenotes}[normal,flushleft]
\footnotesize
\item The metrics, notations, and formatting have the same meanings as those
in Table~\ref{tbl:bpq_ind}.
\end{tablenotes}

\end{threeparttable}
\end{small}
\end{table*}
\begin{table*}[h!]
\centering
\caption{Overall Performance on \ACEP}
\setlength{\tabcolsep}{0pt}%
\label{tbl:acep_ind}
\begin{small}
\begin{threeparttable}

\begin{tabular}{
    @{\hspace{6pt}}l@{\hspace{6pt}}
    @{\hspace{6pt}}r@{\hspace{6pt}}
    @{\hspace{6pt}}r@{\hspace{6pt}}
    @{\hspace{6pt}}r@{\hspace{6pt}}
    @{\hspace{6pt}}r@{\hspace{6pt}}
    @{\hspace{6pt}}r@{\hspace{6pt}}
    @{\hspace{6pt}}r@{\hspace{6pt}}
    @{\hspace{3pt}}r@{\hspace{3pt}}
    @{\hspace{3pt}}r@{\hspace{3pt}}
    @{\hspace{3pt}}r@{\hspace{3pt}}
    @{\hspace{3pt}}r@{\hspace{3pt}}
}
\toprule
\multirow{2}{*}{Model} 
& \multirow{2}{*}{\SR$^{\uparrow}$}
& \multirow{2}{*}{\Val$^{\uparrow}$} 
& \multirow{2}{*}{\Sim$^{\uparrow}$} 
& \multirow{2}{*}{\Nov$^{\uparrow}$}
& \multirow{2}{*}{\SAS$^{\downarrow}$}
& \multirow{2}{*}{\RI$^{\uparrow}$}
& \multicolumn{3}{c}{\APS}
\\
\cmidrule(){8-11} 
& & & & & & & AMP$^\uparrow$ & CARC$^\downarrow$ & hERG$^\downarrow$ & PlogP$^\uparrow$ 
\\
\midrule

\rowcolor{lightgray}
\multicolumn{11}{c}{\textbf{General-purpose LLMs}} 
\\
Mistral (0-shot) & 26.20 & 87.20 & 0.75 & \textbf{\underline{100.00}} & 2.77 & 1.10 & 0.90 & 0.18 & 0.38 & 0.70 \\
Llama (0-shot) & 17.20 & 98.00 & 0.74 & \textbf{\underline{100.00}} & 2.74 & 0.69 & 0.90 & 0.20 & 0.47 & 0.76 \\
Claude-3.5 (0-shot) & 29.60 & 96.20 & 0.71 & \textbf{\underline{100.00}} & 2.78 & 0.69 & 0.91 & 0.17 & 0.38 & 0.64 \\
GPT-4o (0-shot) & 22.20 & 91.40 & 0.74 & 99.10 & 2.77 & 0.52 & 0.90 & 0.17 & \underline{0.36} & 0.54 \\
Mistral (1-shot) & 63.80 & 99.80 & 0.64 & 95.92 & \underline{2.56} & 1.03 & 0.92 & 0.18 & 0.43 & \underline{0.92} \\
Llama (1-shot) & 40.20 & 99.00 & 0.70 & 98.51 & 2.64 & 1.12 & 0.92 & 0.20 & 0.46 & 0.87 \\
Claude-3.5 (1-shot) & 32.60 & 96.60 & 0.71 & \textbf{\underline{100.00}} & 2.74 & \underline{1.24} & \underline{0.94} & \underline{0.16} & 0.42 & 0.60 \\
GPT-4o (1-shot) & 23.00 & 88.80 & \textbf{\underline{0.76}} & \textbf{\underline{100.00}} & 2.79 & 1.09 & 0.93 & 0.17 & 0.40 & 0.63 \\

\rowcolor{lightgray}
\multicolumn{11}{c}{\textbf{Foundational LLMs for Chemistry}} 
\\

\cellcolor{yellow!20}\LlaSMolM  & \underline{\cellcolor{yellow!20}68.60} & \textbf{\underline{\cellcolor{yellow!20}100.00}} & \cellcolor{yellow!20}0.66 & \cellcolor{yellow!20}99.71 & \cellcolor{yellow!20}2.65 & \cellcolor{yellow!20}1.00 & \cellcolor{yellow!20}0.93 & \cellcolor{yellow!20}0.17 & \cellcolor{yellow!20}0.43 & \cellcolor{yellow!20}0.90 \\

\ChemDFML & 22.00 & 93.00 & 0.72 & \textbf{\underline{100.00}} & 2.85 & 1.03 & 0.93 & \underline{0.16} & 0.44 & 0.84 \\

\rowcolor{lightgray}
\multicolumn{11}{c}{\textbf{Specialist LLMs}}
\\

\mollmQuadTaskM  & 85.60 & \textbf{100.00} & 0.54 & 99.53 & 2.39 & \textbf{2.46} & 0.95 & 0.14 & \textbf{0.33} & 1.24 \\

\cellcolor{green!10}\mollmQuadTaskL & \cellcolor{green!10}88.00 & \cellcolor{green!10}99.80 & \cellcolor{green!10}0.54 & \cellcolor{green!10}99.55 & \cellcolor{green!10}2.38 & \cellcolor{green!10}2.24 & \cellcolor{green!10}0.95 & \cellcolor{green!10}0.14 & \cellcolor{green!10}0.34 & \cellcolor{green!10}1.25 \\

\hline
\ImpT & 28.3 & -0.2 & -18.2 & -0.2 & 10.2 & 124.0 & 2.2 & 17.6 & 20.9 & 38.9 \\

\rowcolor{lightgray}
\multicolumn{11}{c}{\textbf{Generalist LLMs}} 
\\

\mollmQuadGenM & 86.60 & \textbf{100.00} & 0.60 & 98.61 & 2.38 & 2.34 & \textbf{0.96} & 0.15 & 0.36 & 1.25 \\

\cellcolor{blue!10}\mollmQuadGenL & \textbf{\cellcolor{blue!10}92.80} & \cellcolor{blue!10}99.80 & \cellcolor{blue!10}0.58 & \cellcolor{blue!10}98.92 & \textbf{\cellcolor{blue!10}2.34} & \cellcolor{blue!10}2.22 & \cellcolor{blue!10}0.95 & \cellcolor{blue!10}0.15 & \cellcolor{blue!10}0.35 & \cellcolor{blue!10}1.26 \\

\mollmDecGenM & 74.60 & \textbf{100.00} & 0.61 & 99.20 & 2.44 & 1.92 & 0.95 & \textbf{0.13} & 0.35 & 1.11 \\
\mollmDecGenL & 72.60 & 93.60 & 0.57 & 98.62 & 2.38 & 2.27 & \textbf{0.96} & 0.15 & 0.38 & \textbf{1.33} \\

\hline
\ImpG & 35.3 & -0.2 & -12.1 & -0.8 & 11.7 & 122.0 & 2.2 & 11.8 & 18.6 & 40.0 \\

\bottomrule
\end{tabular}

\begin{tablenotes}[normal,flushleft]
\footnotesize
\item The metrics, notations, and formatting have the same meanings as those
in Table~\ref{tbl:bpq_ind}.
\end{tablenotes}

\end{threeparttable}
\end{small}
\end{table*}
\begin{table*}[h!]
\centering
\caption{Overall Performance on \BDPQ}
\setlength{\tabcolsep}{0pt}%
\label{tbl:bdpq_ind}
\begin{small}
\begin{threeparttable}

\begin{tabular}{
    @{\hspace{6pt}}l@{\hspace{6pt}}
    @{\hspace{6pt}}r@{\hspace{6pt}}
    @{\hspace{6pt}}r@{\hspace{6pt}}
    @{\hspace{6pt}}r@{\hspace{6pt}}
    @{\hspace{6pt}}r@{\hspace{6pt}}
    @{\hspace{6pt}}r@{\hspace{6pt}}
    @{\hspace{6pt}}r@{\hspace{6pt}}
    @{\hspace{3pt}}r@{\hspace{3pt}}
    @{\hspace{3pt}}r@{\hspace{3pt}}
    @{\hspace{3pt}}r@{\hspace{3pt}}
    @{\hspace{3pt}}r@{\hspace{3pt}}
}
\toprule
\multirow{2}{*}{Model} 
& \multirow{2}{*}{\SR$^{\uparrow}$}
& \multirow{2}{*}{\Val$^{\uparrow}$} 
& \multirow{2}{*}{\Sim$^{\uparrow}$} 
& \multirow{2}{*}{\Nov$^{\uparrow}$}
& \multirow{2}{*}{\SAS$^{\downarrow}$}
& \multirow{2}{*}{\RI$^{\uparrow}$}
& \multicolumn{3}{c}{\APS}
\\
\cmidrule(){8-11} 
& & & & & & & BBBP$^\uparrow$ & DRD2$^\uparrow$ & PlogP$^\uparrow$ & QED$^\uparrow$ 
\\
\midrule

\rowcolor{lightgray}
\multicolumn{11}{c}{\textbf{General-purpose LLMs}} 
\\
Mistral (0-shot) & 2.40 & 75.60 & \textbf{\underline{0.72}} & \textbf{\underline{100.00}} & 2.83 & 0.49 & \textbf{\underline{0.96}} & 0.09 & 0.66 & 0.82 \\
Llama (0-shot) & 8.80 & 97.00 & \textbf{\underline{0.72}} & \textbf{\underline{100.00}} & 3.24 & 1.67 & \textbf{\underline{0.96}} & 0.06 & 0.03 & 0.79 \\
Claude-3.5 (0-shot) & 11.20 & 96.80 & 0.67 & \textbf{\underline{100.00}} & 2.78 & 1.80 & 0.93 & 0.09 & 0.60 & 0.78 \\
GPT-4o (0-shot) & 4.20 & 84.80 & \textbf{\underline{0.72}} & \textbf{\underline{100.00}} & 2.92 & 3.98 & 0.93 & 0.07 & 0.51 & 0.82 \\
Mistral (1-shot) & 21.60 & 99.20 & 0.59 & 92.59 & \underline{2.65} & \underline{4.76} & 0.94 & \underline{0.18} & 0.94 & 0.80 \\
Llama (1-shot) & 14.40 & 99.40 & 0.63 & 91.67 & 3.01 & 2.65 & 0.94 & 0.11 & 0.63 & 0.78 \\
Claude-3.5 (1-shot) & 15.60 & 95.20 & 0.58 & \textbf{\underline{100.00}} & 2.66 & 3.99 & 0.94 & 0.11 & \underline{1.26} & 0.80 \\
GPT-4o (1-shot) & 5.60 & 87.20 & 0.68 & \textbf{\underline{100.00}} & \underline{2.65} & 3.47 & 0.95 & 0.09 & 1.09 & \textbf{\underline{0.85}} \\

\rowcolor{lightgray}
\multicolumn{11}{c}{\textbf{Foundational LLMs for Chemistry}} 
\\

\cellcolor{yellow!20}\LlaSMolM & \underline{\cellcolor{yellow!20}22.60} & \textbf{\underline{\cellcolor{yellow!20}100.00}} & \cellcolor{yellow!20}0.68 & \textbf{\underline{\cellcolor{yellow!20}100.00}} & \cellcolor{yellow!20}2.85 & \cellcolor{yellow!20}2.22 & \cellcolor{yellow!20}0.93 & \cellcolor{yellow!20}0.09 & \cellcolor{yellow!20}0.63 & \cellcolor{yellow!20}0.78 \\

\ChemDFML & 6.20 & 93.00 & 0.67 & \textbf{\underline{100.00}} & 2.85 & 3.51 & 0.92 & 0.07 & 0.64 & 0.80 \\

\rowcolor{lightgray}
\multicolumn{11}{c}{\textbf{Specialist LLMs}}
\\

\cellcolor{green!10}\mollmQuadTaskM & \textbf{\cellcolor{green!10}56.60} & \textbf{\cellcolor{green!10}100.00} & \cellcolor{green!10}0.50 & \cellcolor{green!10}97.88 & \textbf{\cellcolor{green!10}2.45} & \cellcolor{green!10}5.48 & \cellcolor{green!10}0.95 & \textbf{\cellcolor{green!10}0.22} & \cellcolor{green!10}1.25 & \cellcolor{green!10}0.79 \\

\mollmQuadTaskL & 43.60 & 99.80 & 0.58 & 99.08 & 2.52 & 4.85 & 0.95 & 0.16 & 1.14 & 0.79 \\

\hline
\ImpT & 150.4 & 0.0 & -26.5 & -2.1 & 14.0 & 146.8 & 2.2 & 144.4 & 98.4 & 1.3 \\

\rowcolor{lightgray}
\multicolumn{11}{c}{\textbf{Generalist LLMs}} 
\\

\mollmQuadGenM  & 50.60 & \textbf{100.00} & 0.58 & 99.21 & 2.51 & 4.93 & 0.95 & 0.17 & 1.23 & 0.79 \\

\cellcolor{blue!10}\mollmQuadGenL & \cellcolor{blue!10}51.00 & \textbf{\cellcolor{blue!10}100.00} & \cellcolor{blue!10}0.58 & \cellcolor{blue!10}98.43 & \cellcolor{blue!10}2.49 & \cellcolor{blue!10}5.40 & \cellcolor{blue!10}0.95 & \cellcolor{blue!10}0.17 & \cellcolor{blue!10}1.19 & \cellcolor{blue!10}0.78 \\

\mollmDecGenM & 48.40 & 99.40 & 0.58 & 99.17 & 2.55 & 5.05 & 0.95 & 0.16 & 1.22 & 0.79 \\

\mollmDecGenL & 42.60 & 88.60 & 0.55 & 98.59 & 2.47 & \textbf{5.89} & 0.94 & 0.17 & \textbf{1.37} & 0.79 \\

\hline
\ImpG & 125.7 & 0.0 & -14.7 & -1.6 & 12.6 & 143.2 & 2.2 & 88.9 & 88.9 & 0.0 \\

\bottomrule
\end{tabular}

\begin{tablenotes}[normal,flushleft]
\footnotesize
\item The metrics, notations, and formatting have the same meanings as those
in Table~\ref{tbl:bpq_ind}.
\end{tablenotes}

\end{threeparttable}
\end{small}
\end{table*}
\begin{table*}[h!]
\centering
\caption{Overall Performance on \DHMQ}
\setlength{\tabcolsep}{0pt}%
\label{tbl:dhmq_ind}
\begin{small}
\begin{threeparttable}

\begin{tabular}{
    @{\hspace{6pt}}l@{\hspace{6pt}}
    @{\hspace{6pt}}r@{\hspace{6pt}}
    @{\hspace{6pt}}r@{\hspace{6pt}}
    @{\hspace{6pt}}r@{\hspace{6pt}}
    @{\hspace{6pt}}r@{\hspace{6pt}}
    @{\hspace{6pt}}r@{\hspace{6pt}}
    @{\hspace{6pt}}r@{\hspace{6pt}}
    @{\hspace{4pt}}r@{\hspace{4pt}}
    @{\hspace{4pt}}r@{\hspace{4pt}}
    @{\hspace{4pt}}r@{\hspace{4pt}}
    @{\hspace{4pt}}r@{\hspace{4pt}}
}
\toprule
\multirow{2}{*}{Model} 
& \multirow{2}{*}{\SR$^{\uparrow}$}
& \multirow{2}{*}{\Val$^{\uparrow}$} 
& \multirow{2}{*}{\Sim$^{\uparrow}$} 
& \multirow{2}{*}{\Nov$^{\uparrow}$}
& \multirow{2}{*}{\SAS$^{\downarrow}$}
& \multirow{2}{*}{\RI$^{\uparrow}$}
& \multicolumn{4}{c}{\APS}
\\
\cmidrule(){8-11} 
& & & & & & & DRD2$^\uparrow$ & HIA$^\uparrow$ & MUT$^\downarrow$ & QED$^\uparrow$ 
\\
\midrule

\rowcolor{lightgray}
\multicolumn{11}{c}{\textbf{General-purpose LLMs}} 
\\
Mistral (0-shot) & 4.80 & 86.80 & 0.71 & \textbf{\underline{100.00}} & 2.88 & 0.76 & 0.05 & \textbf{\underline{1.00}} & 0.29 & 0.80 \\
Llama (0-shot) & 6.00 & 97.40 & \textbf{\underline{0.73}} & \textbf{\underline{100.00}} & 3.09 & 1.35 & 0.06 & \textbf{\underline{1.00}} & 0.28 & 0.79 \\
Claude-3.5 (0-shot) & 5.20 & 95.20 & 0.63 & \textbf{\underline{100.00}} & \underline{2.73} & 1.84 & 0.10 & \textbf{\underline{1.00}} & 0.20 & 0.75 \\
GPT-4o (0-shot) & 5.80 & 87.80 & 0.72 & \textbf{\underline{100.00}} & 2.89 & 0.88 & 0.07 & \textbf{\underline{1.00}} & 0.22 & \textbf{\underline{0.82}} \\
\cellcolor{yellow!20}Mistral (1-shot) & \underline{\cellcolor{yellow!20}25.60} & \cellcolor{yellow!20}99.80 & \cellcolor{yellow!20}0.55 & \cellcolor{yellow!20}86.72 & \cellcolor{yellow!20}2.89 & \cellcolor{yellow!20}1.89 & \textbf{\underline{\cellcolor{yellow!20}0.18}} & \textbf{\underline{\cellcolor{yellow!20}1.00}} & \cellcolor{yellow!20}0.21 & \cellcolor{yellow!20}0.78 \\
Llama (1-shot) & 13.80 & 99.40 & 0.56 & 85.51 & 3.06 & \textbf{\underline{3.39}} & \textbf{\underline{0.18}} & \textbf{\underline{1.00}} & 0.24 & 0.79 \\
Claude-3.5 (1-shot) & 8.40 & 95.20 & 0.65 & \textbf{\underline{100.00}} & 2.77 & 1.38 & 0.12 & \textbf{\underline{1.00}} & 0.21 & 0.78 \\
GPT-4o (1-shot) & 5.60 & 87.40 & 0.71 & \textbf{\underline{100.00}} & 2.78 & 1.22 & 0.10 & \textbf{\underline{1.00}} & 0.22 & 0.81 \\

\rowcolor{lightgray}
\multicolumn{11}{c}{\textbf{Foundational LLMs for Chemistry}} 
\\

\LlaSMolM & 24.80 & \textbf{\underline{100.00}} & 0.62 & \textbf{\underline{100.00}} & 2.93 & 1.44 & 0.08 & 0.99 & 0.20 & 0.78 \\

\ChemDFML & 6.80 & 86.40 & 0.67 & \textbf{\underline{100.00}} & 3.03 & 1.72 & 0.07 & \textbf{\underline{1.00}} & \textbf{\underline{0.17}} & \textbf{\underline{0.82}} \\

\rowcolor{lightgray}
\multicolumn{11}{c}{\textbf{Specialist LLMs}}
\\

\cellcolor{green!10}\mollmQuadTaskM  & \cellcolor{green!10}44.60 & \cellcolor{green!10}99.80 & \cellcolor{green!10}0.57 & \cellcolor{green!10}99.10 & \cellcolor{green!10}2.81 & \cellcolor{green!10}2.96 & \cellcolor{green!10}0.14 & \cellcolor{green!10}0.99 & \cellcolor{green!10}0.19 & \cellcolor{green!10}0.78 \\

\mollmQuadTaskL  & 35.40 & \textbf{100.00} & 0.65 & \textbf{100.00} & 2.73 & 2.63 & 0.12 & 0.99 & 0.20 & 0.79 \\

\hline
\ImpT & 74.2 & 0.0 & 3.6 & 14.3 & 2.8 & 56.6 & -22.2 & -1.0 & 9.5 & 0.0 \\

\rowcolor{lightgray}
\multicolumn{11}{c}{\textbf{Generalist LLMs}} 
\\

\cellcolor{blue!10}\mollmQuadGenM & \textbf{\cellcolor{blue!10}53.40} & \textbf{\cellcolor{blue!10}100.00} & \cellcolor{blue!10}0.59 & \cellcolor{blue!10}99.25 & \cellcolor{blue!10}2.76 & \cellcolor{blue!10}3.26 & \cellcolor{blue!10}0.15 & \cellcolor{blue!10}0.99 & \cellcolor{blue!10}0.19 & \cellcolor{blue!10}0.78 \\

\mollmQuadGenL & 50.40 & \textbf{100.00} & 0.59 & \textbf{100.00} & 2.67 & 3.28 & 0.13 & 0.99 & 0.19 & 0.79 \\

\mollmDecGenM  & 52.20 & 99.60 & 0.61 & \textbf{100.00} & 2.76 & 2.24 & 0.12 & 0.99 & 0.19 & 0.79 \\
\mollmDecGenL & 41.80 & 83.20 & 0.57 & \textbf{100.00} & \textbf{2.65} & 3.32 & 0.15 & 0.99 & 0.20 & 0.79 \\

\hline
\ImpG & 108.6 & 0.2 & 7.3 & 14.4 & 4.5 & 72.5 & -16.7 & -1.0 & 9.5 & 0.0 \\

\bottomrule
\end{tabular}

\begin{tablenotes}[normal,flushleft]
\footnotesize
\item The metrics, notations, and formatting have the same meanings as those
in Table~\ref{tbl:bpq_ind}.
\end{tablenotes}

\end{threeparttable}
\end{small}
\end{table*}

\subsection{OOD Evaluation}
\label{sec:app:results:ood}

Tables~\ref{tbl:cde_ood},~\ref{tbl:abmp_ood},~\ref{tbl:bcmq_ood},~\ref{tbl:bdeq_ood} and \ref{tbl:hlmpq_ood} presents the performance comparison of
{\mollm}s with general-purpose LLMs and \LlaSMolM under all evaluation metrics for each OOD task.

\begin{table*}[h!]
\centering
\caption{Overall Performance on \CDE}
\setlength{\tabcolsep}{0pt}%
\label{tbl:cde_ood}
\begin{small}
\begin{threeparttable}

\begin{tabular}{
    @{\hspace{9pt}}l@{\hspace{9pt}}
    @{\hspace{9pt}}r@{\hspace{9pt}}
    @{\hspace{9pt}}r@{\hspace{9pt}}
    @{\hspace{9pt}}r@{\hspace{9pt}}
    @{\hspace{9pt}}r@{\hspace{9pt}}
    @{\hspace{9pt}}r@{\hspace{9pt}}
    @{\hspace{9pt}}r@{\hspace{9pt}}
    @{\hspace{4pt}}r@{\hspace{4pt}}
    @{\hspace{4pt}}r@{\hspace{4pt}}
    @{\hspace{4pt}}r@{\hspace{4pt}}
}
\toprule
\multirow{2}{*}{Model} 
& \multirow{2}{*}{\SR$^{\uparrow}$}
& \multirow{2}{*}{\Val$^{\uparrow}$} 
& \multirow{2}{*}{\Sim$^{\uparrow}$} 
& \multirow{2}{*}{\Nov$^{\uparrow}$}
& \multirow{2}{*}{\SAS$^{\downarrow}$}
& \multirow{2}{*}{\RI$^{\uparrow}$}
& \multicolumn{3}{c}{\APS}
\\
\cmidrule(){8-10} 
& & & & & & & CARC$^\downarrow$ & DRD2$^\uparrow$ & hERG$^\downarrow$ 
\\
\midrule

\rowcolor{lightgray}
\multicolumn{10}{c}{\textbf{General-purpose LLMs}} 
\\
Mistral (0-shot) & 3.00 & 86.00 & 0.73 & \textbf{\underline{100.00}} & 3.13 & 1.33 & 0.15 & \textbf{\underline{0.14}} & 0.65 \\
Llama (0-shot) & 6.80 & 96.60 & 0.68 & \textbf{\underline{100.00}} & 3.32 & 0.77 & 0.20 & 0.06 & 0.57 \\
Claude-3.5 (0-shot) & 6.80 & 97.80 & 0.70 & \textbf{\underline{100.00}} & 2.98 & 1.07 & 0.16 & 0.08 & 0.52 \\
GPT-4o (0-shot) & 3.80 & 89.80 & \textbf{\underline{0.74}} & \textbf{\underline{100.00}} & 3.01 & 1.56 & 0.15 & 0.05 & \textbf{\underline{0.39}} \\
\cellcolor{yellow!20}Mistral (1-shot) & \underline{\cellcolor{yellow!20}30.60} & \cellcolor{yellow!20}99.60 & \cellcolor{yellow!20}0.62 & \cellcolor{yellow!20}93.46 & \cellcolor{yellow!20}3.00 & \textbf{\underline{\cellcolor{yellow!20}1.66}} & \cellcolor{yellow!20}0.15 & \cellcolor{yellow!20}0.09 & \cellcolor{yellow!20}0.50 \\
Llama (1-shot) & 18.20 & 99.40 & 0.55 & 76.92 & 3.50 & 1.51 & 0.14 & 0.12 & 0.47 \\
Claude-3.5 (1-shot) & 8.40 & 98.40 & 0.66 & \textbf{\underline{100.00}} & 2.91 & 1.09 & \underline{0.12} & 0.08 & 0.47 \\
GPT-4o (1-shot) & 7.00 & 88.20 & 0.72 & \textbf{\underline{100.00}} & 3.10 & 1.04 & 0.16 & 0.05 & 0.53 \\

\rowcolor{lightgray}
\multicolumn{10}{c}{\textbf{Foundational LLMs for Chemistry}} 
\\

\LlaSMolM & 29.80 & \textbf{\underline{99.80}} & 0.61 & 97.99 & \textbf{\underline{2.79}} & 1.28 & 0.14 & 0.06 & 0.46 \\
\ChemDFML & 8.20 & 90.60 & 0.64 & \textbf{\underline{100.00}} & 3.16 & 0.84 & 0.17 & 0.08 & 0.53 \\

\rowcolor{lightgray}
\multicolumn{10}{c}{\textbf{Generalist LLMs}} 
\\

\cellcolor{blue!10}\mollmDecGenM 
 & \textbf{\cellcolor{blue!10}39.80} & \cellcolor{blue!10}98.60 & \cellcolor{blue!10}0.58 & \textbf{\cellcolor{blue!10}100.00} & \cellcolor{blue!10}2.85 & \textbf{\cellcolor{blue!10}1.66} & \textbf{\cellcolor{blue!10}0.11} & \cellcolor{blue!10}0.08 & \cellcolor{blue!10}0.42 \\

\mollmDecGenL 
& 33.20 & 86.80 & 0.55 & \textbf{100.00} & 2.86 & 1.50 & \textbf{0.11} & 0.08 & 0.48 \\

\hline
\ImpG 
 & 30.1 & -1.0 & -6.5 & 7.0 & 5.0 & 0.0 & 26.7 & -11.1 & 16.0 \\

\bottomrule
\end{tabular}

\begin{tablenotes}[normal,flushleft]
\footnotesize
\item The metrics, notations, and formatting have the same meanings as those
in Table~\ref{tbl:bpq_ind}.
\end{tablenotes}

\end{threeparttable}
\end{small}
\end{table*}
\begin{table*}[h!]
\centering
\caption{Overall Performance on \ABMP}
\setlength{\tabcolsep}{0pt}%
\label{tbl:abmp_ood}
\begin{small}
\begin{threeparttable}

\begin{tabular}{
    @{\hspace{6pt}}l@{\hspace{6pt}}
    @{\hspace{6pt}}r@{\hspace{6pt}}
    @{\hspace{6pt}}r@{\hspace{6pt}}
    @{\hspace{6pt}}r@{\hspace{6pt}}
    @{\hspace{6pt}}r@{\hspace{6pt}}
    @{\hspace{6pt}}r@{\hspace{6pt}}
    @{\hspace{6pt}}r@{\hspace{6pt}}
    @{\hspace{4pt}}r@{\hspace{4pt}}
    @{\hspace{4pt}}r@{\hspace{4pt}}
    @{\hspace{4pt}}r@{\hspace{4pt}}
    @{\hspace{4pt}}r@{\hspace{4pt}}
}
\toprule
\multirow{2}{*}{Model} 
& \multirow{2}{*}{\SR$^{\uparrow}$}
& \multirow{2}{*}{\Val$^{\uparrow}$} 
& \multirow{2}{*}{\Sim$^{\uparrow}$} 
& \multirow{2}{*}{\Nov$^{\uparrow}$}
& \multirow{2}{*}{\SAS$^{\downarrow}$}
& \multirow{2}{*}{\RI$^{\uparrow}$}
& \multicolumn{3}{c}{\APS}
\\
\cmidrule(){8-11} 
& & & & & & & AMP$^\uparrow$ & BBBP$^\uparrow$ & MUT$^\downarrow$ & PlogP$^\uparrow$ 
\\
\midrule

\rowcolor{lightgray}
\multicolumn{11}{c}{\textbf{General-purpose LLMs}} 
\\
Mistral (0-shot) & 23.00 & 83.00 & \textbf{\underline{0.77}} & \textbf{\underline{100.00}} & 2.76 & 0.93 & 0.90 & 0.87 & 0.24 & 0.86 \\
Llama (0-shot) & 44.60 & 98.40 & 0.71 & \textbf{\underline{100.00}} & 2.85 & 0.61 & 0.92 & 0.90 & 0.25 & 1.17 \\
Claude-3.5 (0-shot) & 43.60 & 96.20 & 0.70 & \textbf{\underline{100.00}} & 2.73 & 0.80 & \underline{0.95} & 0.89 & 0.24 & 0.81 \\
GPT-4o (0-shot) & 27.00 & 87.40 & 0.73 & \textbf{\underline{100.00}} & 2.72 & 0.51 & 0.93 & 0.89 & 0.25 & 0.93 \\
\cellcolor{yellow!20}Mistral (1-shot) & \underline{\cellcolor{yellow!20}73.20} & \cellcolor{yellow!20}99.60 & \cellcolor{yellow!20}0.64 & \cellcolor{yellow!20}94.81 & \underline{\cellcolor{yellow!20}2.62} & \underline{\cellcolor{yellow!20}1.09} & \cellcolor{yellow!20}0.93 & \cellcolor{yellow!20}0.90 & \underline{\cellcolor{yellow!20}0.23} & \cellcolor{yellow!20}1.10 \\
Llama (1-shot) & 60.80 & 99.60 & 0.70 & 99.01 & 2.76 & 0.83 & 0.92 & 0.89 & 0.24 & 1.02 \\
Claude-3.5 (1-shot) & 45.20 & 96.40 & 0.64 & \textbf{\underline{100.00}} & 2.67 & 0.87 & \underline{0.95} & \underline{0.91} & \underline{0.23} & 1.04 \\
GPT-4o (1-shot) & 34.40 & 87.80 & 0.74 & \textbf{\underline{100.00}} & 2.73 & 0.65 & 0.93 & 0.89 & 0.28 & 1.03 \\

\rowcolor{lightgray}
\multicolumn{11}{c}{\textbf{Foundational LLMs for Chemistry}} 
\\

\LlaSMolM & 72.40 & \textbf{\underline{100.00}} & 0.67 & \textbf{\underline{100.00}} & 2.75 & 0.78 & 0.94 & 0.89 & 0.24 & 0.93 \\

\ChemDFML & 39.60 & 92.40 & 0.67 & \textbf{\underline{100.00}} & 2.95 & 0.98 & 0.94 & 0.89 & \underline{0.23} & \underline{1.40} \\

\rowcolor{lightgray}
\multicolumn{11}{c}{\textbf{Generalist LLMs}} 
\\

\cellcolor{blue!10}\mollmDecGenM & \textbf{\cellcolor{blue!10}86.60} & \cellcolor{blue!10}99.40 & \cellcolor{blue!10}0.63 & \cellcolor{blue!10}98.85 & \cellcolor{blue!10}2.48 & \cellcolor{blue!10}1.68 & \cellcolor{blue!10}0.95 & \textbf{\cellcolor{blue!10}0.92} & \cellcolor{blue!10}0.20 & \cellcolor{blue!10}1.63 \\

\mollmDecGenL  & 79.60 & 89.60 & 0.58 & 98.99 & \textbf{2.42} & \textbf{1.81} & \textbf{0.96} & 0.91 & \textbf{0.19} & \textbf{1.81} \\
\hline
\ImpG & 18.3 & -0.2 & -1.6 & 4.3 & 5.3 & 54.1 & 2.2 & 2.2 & 13.0 & 48.2 \\

\bottomrule
\end{tabular}

\begin{tablenotes}[normal,flushleft]
\footnotesize
\item The metrics, notations, and formatting have the same meanings as those
in Table~\ref{tbl:bpq_ind}.
\end{tablenotes}

\end{threeparttable}
\end{small}
\end{table*}
\begin{table*}[h!]
\centering
\caption{Overall Performance on \BCMQ}
\setlength{\tabcolsep}{0pt}%
\label{tbl:bcmq_ood}
\begin{small}
\begin{threeparttable}

\begin{tabular}{
    @{\hspace{6pt}}l@{\hspace{6pt}}
    @{\hspace{6pt}}r@{\hspace{6pt}}
    @{\hspace{6pt}}r@{\hspace{6pt}}
    @{\hspace{6pt}}r@{\hspace{6pt}}
    @{\hspace{6pt}}r@{\hspace{6pt}}
    @{\hspace{6pt}}r@{\hspace{6pt}}
    @{\hspace{6pt}}r@{\hspace{6pt}}
    @{\hspace{4pt}}r@{\hspace{4pt}}
    @{\hspace{4pt}}r@{\hspace{4pt}}
    @{\hspace{4pt}}r@{\hspace{4pt}}
    @{\hspace{4pt}}r@{\hspace{4pt}}
}
\toprule
\multirow{2}{*}{Model} 
& \multirow{2}{*}{\SR$^{\uparrow}$}
& \multirow{2}{*}{\Val$^{\uparrow}$} 
& \multirow{2}{*}{\Sim$^{\uparrow}$} 
& \multirow{2}{*}{\Nov$^{\uparrow}$}
& \multirow{2}{*}{\SAS$^{\downarrow}$}
& \multirow{2}{*}{\RI$^{\uparrow}$}
& \multicolumn{3}{c}{\APS}
\\
\cmidrule(){8-11} 
& & & & & & & BBBP$^\uparrow$ & CARC$^\downarrow$ & MUT$^\downarrow$ & QED$^\uparrow$ 
\\
\midrule

\rowcolor{lightgray}
\multicolumn{11}{c}{\textbf{General-purpose LLMs}} 
\\
Mistral (0-shot) & 25.40 & 89.60 & 0.69 & \textbf{\underline{100.00}} & 2.84 & 0.25 & \underline{0.92} & 0.16 & 0.25 & 0.77 \\
Llama (0-shot) & 20.40 & 98.60 & 0.72 & \textbf{\underline{100.00}} & 2.86 & 0.20 & 0.90 & 0.18 & 0.24 & \underline{0.79} \\
Claude-3.5 (0-shot) & 30.00 & 96.00 & 0.64 & \textbf{\underline{100.00}} & 2.66 & 0.26 & 0.91 & 0.16 & 0.22 & 0.77 \\
GPT-4o (0-shot) & 19.60 & 90.60 & 0.72 & \textbf{\underline{100.00}} & 2.66 & 0.19 & 0.90 & 0.18 & 0.21 & 0.77 \\
Mistral (1-shot) & 63.80 & 99.60 & 0.60 & 93.10 & \underline{2.61} & \underline{0.31} & 0.90 & 0.16 & \underline{0.20} & 0.78 \\
Llama (1-shot) & 41.60 & 99.80 & 0.67 & 95.67 & 2.78 & 0.23 & 0.91 & 0.17 & 0.23 & 0.77 \\
Claude-3.5 (1-shot) & 32.40 & 95.00 & 0.61 & \textbf{\underline{100.00}} & 2.69 & 0.30 & 0.91 & 0.15 & 0.23 & 0.78 \\
GPT-4o (1-shot) & 23.40 & 86.40 & \textbf{\underline{0.73}} & \textbf{\underline{100.00}} & 2.63 & 0.21 & 0.90 & 0.18 & \underline{0.20} & 0.76 \\

\rowcolor{lightgray}
\multicolumn{11}{c}{\textbf{Foundational LLMs for Chemistry}} 
\\

\cellcolor{yellow!20}\LlaSMolM & \underline{\cellcolor{yellow!20}72.80} & \textbf{\underline{\cellcolor{yellow!20}100.00}} & \cellcolor{yellow!20}0.63 & \cellcolor{yellow!20}98.90 & \cellcolor{yellow!20}2.71 & \cellcolor{yellow!20}0.30 & \cellcolor{yellow!20}0.90 & \cellcolor{yellow!20}0.16 & \underline{\cellcolor{yellow!20}0.20} & \cellcolor{yellow!20}0.77 \\

\ChemDFML & 18.20 & 87.00 & 0.67 & 98.90 & 2.90 & 0.27 & 0.90 & \underline{0.14} & 0.23 & 0.76 \\

\rowcolor{lightgray}
\multicolumn{11}{c}{\textbf{Generalist LLMs}} 
\\

\cellcolor{blue!10}\mollmDecGenM  & \textbf{\cellcolor{blue!10}84.20} & \cellcolor{blue!10}99.20 & \cellcolor{blue!10}0.62 & \cellcolor{blue!10}99.52 & \cellcolor{blue!10}2.55 & \cellcolor{blue!10}0.42 & \textbf{\cellcolor{blue!10}0.93} & \textbf{\cellcolor{blue!10}0.12} & \textbf{\cellcolor{blue!10}0.17} & \cellcolor{blue!10}0.81 \\

\mollmDecGenL  & 80.00 & 91.20 & 0.57 & 99.00 & \textbf{2.49} & \textbf{0.44} & \textbf{0.93} & \textbf{0.12} & \textbf{0.17} & \textbf{0.82} \\

\hline
\ImpG  & 15.7 & -0.8 & -1.6 & 0.6 & 5.9 & 40.0 & 3.3 & 25.0 & 15.0 & 5.2 \\

\bottomrule
\end{tabular}

\begin{tablenotes}[normal,flushleft]
\footnotesize
\item The metrics, notations, and formatting have the same meanings as those
in Table~\ref{tbl:bpq_ind}.
\end{tablenotes}

\end{threeparttable}
\end{small}
\end{table*}
\begin{table*}[h!]
\centering
\caption{Overall Performance on \BDEQ}
\setlength{\tabcolsep}{0pt}%
\label{tbl:bdeq_ood}
\begin{small}
\begin{threeparttable}

\begin{tabular}{
    @{\hspace{6pt}}l@{\hspace{6pt}}
    @{\hspace{6pt}}r@{\hspace{6pt}}
    @{\hspace{6pt}}r@{\hspace{6pt}}
    @{\hspace{6pt}}r@{\hspace{6pt}}
    @{\hspace{6pt}}r@{\hspace{6pt}}
    @{\hspace{6pt}}r@{\hspace{6pt}}
    @{\hspace{6pt}}r@{\hspace{6pt}}
    @{\hspace{4pt}}r@{\hspace{4pt}}
    @{\hspace{4pt}}r@{\hspace{4pt}}
    @{\hspace{4pt}}r@{\hspace{4pt}}
    @{\hspace{4pt}}r@{\hspace{4pt}}
}
\toprule
\multirow{2}{*}{Model} 
& \multirow{2}{*}{\SR$^{\uparrow}$}
& \multirow{2}{*}{\Val$^{\uparrow}$} 
& \multirow{2}{*}{\Sim$^{\uparrow}$} 
& \multirow{2}{*}{\Nov$^{\uparrow}$}
& \multirow{2}{*}{\SAS$^{\downarrow}$}
& \multirow{2}{*}{\RI$^{\uparrow}$}
& \multicolumn{3}{c}{\APS}
\\
\cmidrule(){8-11} 
& & & & & & & BBBP$^\uparrow$ & DRD2$^\uparrow$ & hERG$^\downarrow$ & QED$^\uparrow$ 
\\
\midrule

\rowcolor{lightgray}
\multicolumn{11}{c}{\textbf{General-purpose LLMs}} 
\\
Mistral (0-shot) & 3.00 & 78.00 & \textbf{\underline{0.71}} & \textbf{\underline{100.00}} & 2.97 & 1.05 & 0.88 & 0.06 & \textbf{\underline{0.40}} & 0.75 \\
Llama (0-shot) & 2.20 & 96.00 & 0.68 & \textbf{\underline{100.00}} & 3.46 & 0.60 & \textbf{\underline{0.96}} & 0.07 & 0.48 & 0.78 \\
Claude-3.5 (0-shot) & 4.80 & 96.60 & 0.62 & \textbf{\underline{100.00}} & 2.76 & 0.57 & 0.92 & 0.04 & 0.52 & 0.79 \\
GPT-4o (0-shot) & 3.40 & 87.60 & \textbf{\underline{0.71}} & \textbf{\underline{100.00}} & \textbf{\underline{2.75}} & 0.42 & 0.93 & 0.07 & 0.55 & \textbf{\underline{0.82}} \\
\cellcolor{yellow!20}Mistral (1-shot) & \underline{\cellcolor{yellow!20}21.60} & \cellcolor{yellow!20}99.80 & \cellcolor{yellow!20}0.58 & \cellcolor{yellow!20}84.26 & \cellcolor{yellow!20}3.11 & \cellcolor{yellow!20}1.16 & \cellcolor{yellow!20}0.91 & \cellcolor{yellow!20}0.15 & \cellcolor{yellow!20}0.49 & \cellcolor{yellow!20}0.77 \\
Llama (1-shot) & 11.40 & 99.60 & 0.51 & 68.42 & 3.48 & 1.54 & 0.92 & \textbf{\underline{0.19}} & 0.49 & 0.79 \\
Claude-3.5 (1-shot) & 7.20 & 97.60 & 0.55 & \textbf{\underline{100.00}} & 2.88 & 1.22 & 0.95 & 0.08 & 0.53 & 0.79 \\
GPT-4o (1-shot) & 2.20 & 86.00 & 0.70 & \textbf{\underline{100.00}} & 2.81 & 0.83 & 0.95 & 0.09 & 0.57 & 0.80 \\

\rowcolor{lightgray}
\multicolumn{11}{c}{\textbf{Foundational LLMs for Chemistry}} 
\\

\LlaSMolM & 18.20 & \textbf{\underline{100.00}} & 0.60 & \textbf{\underline{100.00}} & 2.86 & 0.65 & 0.92 & 0.07 & 0.49 & 0.80 \\
\ChemDFML & 3.00 & 87.40 & 0.68 & \textbf{\underline{100.00}} & 3.13 & \textbf{\underline{1.64}} & 0.94 & 0.08 & 0.49 & 0.79 \\

\rowcolor{lightgray}
\multicolumn{11}{c}{\textbf{Generalist LLMs}} 
\\

\cellcolor{blue!10}\mollmDecGenM & \textbf{\cellcolor{blue!10}29.20} & \cellcolor{blue!10}98.40 & \cellcolor{blue!10}0.60 & \textbf{\cellcolor{blue!10}100.00} & \cellcolor{blue!10}2.78 & \cellcolor{blue!10}1.22 & \cellcolor{blue!10}0.92 & \cellcolor{blue!10}0.08 & \cellcolor{blue!10}0.45 & \cellcolor{blue!10}0.80 \\

\mollmDecGenL & 28.40 & 92.20 & 0.58 & \textbf{100.00} & \textbf{2.75} & 0.88 & 0.92 & 0.07 & 0.47 & 0.80 \\
\hline
\ImpG & 35.2 & -1.4 & 3.4 & 18.7 & 10.6 & 5.2 & 1.1 & -46.7 & 8.2 & 3.9 \\

\bottomrule
\end{tabular}

\begin{tablenotes}[normal,flushleft]
\footnotesize
\item The metrics, notations, and formatting have the same meanings as those
in Table~\ref{tbl:bpq_ind}.
\end{tablenotes}

\end{threeparttable}
\end{small}
\end{table*}
\begin{table*}[h!]
\centering
\caption{Overall Performance on \HLMPQ}
\setlength{\tabcolsep}{0pt}%
\label{tbl:hlmpq_ood}
\begin{small}
\begin{threeparttable}

\begin{tabular}{
    @{\hspace{6pt}}l@{\hspace{6pt}}
    @{\hspace{6pt}}r@{\hspace{6pt}}
    @{\hspace{6pt}}r@{\hspace{6pt}}
    @{\hspace{6pt}}r@{\hspace{6pt}}
    @{\hspace{6pt}}r@{\hspace{6pt}}
    @{\hspace{6pt}}r@{\hspace{6pt}}
    @{\hspace{6pt}}r@{\hspace{6pt}}
    @{\hspace{3pt}}r@{\hspace{3pt}}
    @{\hspace{3pt}}r@{\hspace{3pt}}
    @{\hspace{3pt}}r@{\hspace{3pt}}
    @{\hspace{3pt}}r@{\hspace{3pt}}
    @{\hspace{3pt}}r@{\hspace{3pt}}
}
\toprule
\multirow{2}{*}{Model} 
& \multirow{2}{*}{\SR$^{\uparrow}$}
& \multirow{2}{*}{\Val$^{\uparrow}$} 
& \multirow{2}{*}{\Sim$^{\uparrow}$} 
& \multirow{2}{*}{\Nov$^{\uparrow}$}
& \multirow{2}{*}{\SAS$^{\downarrow}$}
& \multirow{2}{*}{\RI$^{\uparrow}$}
& \multicolumn{3}{c}{\APS}
\\
\cmidrule(){8-12} 
& & & & & & & HIA$^\uparrow$ & LIV$^\downarrow$ & MUT$^\downarrow$ 
& PlogP$^\uparrow$ & QED$^\uparrow$ 
\\
\midrule

\rowcolor{lightgray}
\multicolumn{12}{c}{\textbf{General-purpose LLMs}} 
\\
Mistral (0-shot) & 11.60 & 82.40 & \textbf{\underline{0.79}} & \textbf{\underline{100.00}} & 2.91 & \textbf{\underline{1.76}} & 0.99 & \textbf{\underline{0.38}} & 0.20 & 0.51 & 0.77 \\
Llama (0-shot) & 20.20 & 99.40 & 0.72 & 98.02 & 2.82 & 0.68 & \textbf{\underline{1.00}} & 0.54 & 0.23 & 0.70 & \textbf{\underline{0.79}} \\
Claude-3.5 (0-shot) & 21.00 & 97.00 & 0.66 & 99.05 & 2.72 & 0.59 & \textbf{\underline{1.00}} & 0.46 & 0.24 & 0.69 & \textbf{\underline{0.79}} \\
GPT-4o (0-shot) & 12.80 & 87.60 & 0.72 & \textbf{\underline{100.00}} & 2.78 & 0.47 & \textbf{\underline{1.00}} & 0.48 & 0.20 & 0.49 & 0.75 \\
\cellcolor{yellow!20}Mistral (1-shot) & \underline{\cellcolor{yellow!20}55.60} & \cellcolor{yellow!20}99.80 & \cellcolor{yellow!20}0.62 & \cellcolor{yellow!20}97.12 & \underline{\cellcolor{yellow!20}2.59} & \cellcolor{yellow!20}0.77 & \cellcolor{yellow!20}0.99 & \cellcolor{yellow!20}0.54 & \cellcolor{yellow!20}0.21 & \underline{\cellcolor{yellow!20}1.08} & \cellcolor{yellow!20}0.77 \\
Llama (1-shot) & 28.00 & 99.60 & 0.70 & 97.86 & 2.72 & 0.75 & \textbf{\underline{1.00}} & 0.56 & 0.24 & 0.83 & 0.78 \\
Claude-3.5 (1-shot) & 25.00 & 95.00 & 0.61 & 97.60 & 2.60 & 0.72 & \textbf{\underline{1.00}} & 0.53 & 0.25 & 0.89 & 0.78 \\
GPT-4o (1-shot) & 13.40 & 87.40 & 0.71 & \textbf{\underline{100.00}} & 2.82 & 0.65 & \textbf{\underline{1.00}} & 0.50 & 0.21 & 0.61 & 0.73 \\

\rowcolor{lightgray}
\multicolumn{12}{c}{\textbf{Foundational LLMs for Chemistry}} 
\\

\LlaSMolM & 37.80 & \textbf{\underline{100.00}} & 0.68 & \textbf{\underline{100.00}} & 2.66 & 0.66 & \textbf{\underline{1.00}} & 0.58 & 0.22 & 0.92 & 0.73 \\

\ChemDFML & 10.80 & 90.60 & 0.68 & 98.15 & 3.01 & 1.04 & 0.98 & 0.43 & \underline{0.19} & 0.68 & 0.77 \\

\rowcolor{lightgray}
\multicolumn{12}{c}{\textbf{Generalist LLMs}} 
\\

\cellcolor{blue!10}\mollmDecGenM & \textbf{\cellcolor{blue!10}74.60} & \cellcolor{blue!10}99.80 & \cellcolor{blue!10}0.61 & \cellcolor{blue!10}99.46 & \cellcolor{blue!10}2.49 & \cellcolor{blue!10}1.36 & \textbf{\cellcolor{blue!10}1.00} & \cellcolor{blue!10}0.53 & \textbf{\cellcolor{blue!10}0.18} & \cellcolor{blue!10}1.43 & \textbf{\cellcolor{blue!10}0.79} \\

\mollmDecGenL & 65.40 & 90.80 & 0.58 & 99.69 & \textbf{2.41} & 1.35 & \textbf{1.00} & 0.53 & \textbf{0.18} & \textbf{1.53} & \textbf{0.79} \\
\hline
\ImpG  & 34.2 & 0.0 & -1.6 & 2.4 & 3.9 & 76.6 & 1.0 & 1.9 & 14.3 & 32.4 & 2.6 \\

\bottomrule
\end{tabular}

\begin{tablenotes}[normal,flushleft]
\footnotesize
\item The metrics, notations, and formatting have the same meanings as those
in Table~\ref{tbl:bpq_ind}.
\end{tablenotes}

\end{threeparttable}
\end{small}
\end{table*}

\subsection{IND Evaluation with Unseen Instructions}
\label{sec:app:results:uninst}

Table~\ref{tbl:main_uninst} presents the overall performance comparison
of specialist and generalist {\mollm}s when evaluated with seen and unseen instructions.

\begin{table*}[h!]
\centering
\setlength{\tabcolsep}{0pt}%
\caption{Overall Performance with Unseen Instructions in IND Tasks}
\label{tbl:main_uninst}
\vspace{-5pt}
\begin{small}
\begin{threeparttable}
\begin{tabular}{
   @{\hspace{2pt}}l@{\hspace{2pt}}
   @{\hspace{2pt}}l@{\hspace{10pt}}
   @{\hspace{2pt}}r@{\hspace{2pt}}
   @{\hspace{2pt}}r@{\hspace{2pt}}
   @{\hspace{2pt}}r@{\hspace{2pt}}
   @{\hspace{5pt}}c@{\hspace{5pt}}
   @{\hspace{0pt}}r@{\hspace{2pt}}
   @{\hspace{2pt}}r@{\hspace{2pt}}
   @{\hspace{2pt}}r@{\hspace{2pt}}
   @{\hspace{5pt}}c@{\hspace{5pt}}
   @{\hspace{2pt}}r@{\hspace{2pt}}
   @{\hspace{2pt}}r@{\hspace{2pt}}
   @{\hspace{2pt}}r@{\hspace{2pt}}
   @{\hspace{5pt}}c@{\hspace{5pt}}
   @{\hspace{0pt}}r@{\hspace{2pt}}
   @{\hspace{2pt}}r@{\hspace{2pt}}
   @{\hspace{2pt}}r@{\hspace{2pt}}
   @{\hspace{5pt}}c@{\hspace{5pt}}
   @{\hspace{0pt}}r@{\hspace{2pt}}
   @{\hspace{2pt}}r@{\hspace{2pt}}
   @{\hspace{2pt}}r@{\hspace{0pt}}
}
\toprule
\multirow{2}{*}{Model} &
\multirow{2}{*}{Instr} & 
\multicolumn{3}{c}{\BPQ} && \multicolumn{3}{c}{\ELQ} && \multicolumn{3}{c}{\ACEP} && \multicolumn{3}{c}{\BDPQ} && \multicolumn{3}{c}{\DHMQ} \\
\cmidrule(){3-5} \cmidrule(){7-9} \cmidrule(){11-13} \cmidrule(){15-17} \cmidrule(){19-21}
\mollm &  & \SR$^{\uparrow}$ & \Sim$^{\uparrow}$ & \RI$^{\uparrow}$ &
& \SR$^{\uparrow}$ & \Sim$^{\uparrow}$ & \RI$^{\uparrow}$ &
& \SR$^{\uparrow}$ & \Sim$^{\uparrow}$ & \RI$^{\uparrow}$ &
& \SR$^{\uparrow}$ & \Sim$^{\uparrow}$ & \RI$^{\uparrow}$ &
& \SR$^{\uparrow}$ & \Sim$^{\uparrow}$ & \RI$^{\uparrow}$ \\
\midrule

\rowcolor{lightgray}
\multicolumn{21}{c}{\textbf{Specialist LLMs}}
\\
\multirow{2}{*}{{\mbox{$\mathop{\mathtt{\text{-}N_{Mistral}}}\limits$}\xspace}}
& seen 
 & 71.00 & 0.57 & \textbf{2.59} &
 & 81.80 & 0.55 & 0.39 &
 & 85.60 & 0.54 & \textbf{2.46} &
& 56.60 & \textbf{0.50} & 5.48 &
& 44.60 & 0.57 & 2.96 
\\
& unseen 
& 68.60 & 0.55 & 2.33 &
& 84.60 & 0.53 & \textbf{0.41} &
& 86.80 & 0.53 & 2.28 &
& 59.40 & 0.47 & \textbf{5.79} &
& \textbf{49.40} & 0.56 & \textbf{3.19} 
\\
\hline

\multirow{2}{*}{{\mbox{$\mathop{\mathtt{\text{-}N_{Llama}}}\limits$}\xspace}}
& seen 
& \textbf{84.20} & 0.58 & 2.09 &
 & 85.40 & 0.53 & 0.41 & 
 & 88.00 & 0.54 & 2.24 & 
 & \textbf{43.60} & 0.58 & 4.85 &
 & 35.40 & 0.65 & 2.63 
\\

& unseen 
& 74.20 & 0.57 & 2.02 &
& 88.60 & 0.54 & 0.42 &
& 87.00 & 0.52 & 2.14 &
& 37.00 & 0.59 & \textbf{5.27} &
& \textbf{37.60} & 0.64 & \textbf{2.77}
\\

\rowcolor{lightgray}
\multicolumn{21}{c}{\textbf{Generalist LLMs}}
\\

\multirow{2}{*}{{\mbox{$\mathop{\mathtt{\text{-}P(10)_{Mistral}}}\limits$}\xspace}} 
& seen 
& 89.40 & 0.62 & \textbf{2.30} &  & 88.40 & 0.59 & \textbf{0.41} &  & 74.60 & 0.61 & \textbf{1.92} &  & 48.40 & 0.58 & \textbf{5.05} &  & 52.20 & 0.61 & 2.24 
\\
 & unseen 
 & 89.60 & 0.62 & 2.01 &  & 87.60 & 0.60 & 0.37 &  & 78.00 & 0.63 & 1.75 &  & 46.60 & 0.60 & 4.57 &  & 50.20 & 0.61 & \textbf{2.79}
 \\
 
\hline
\multirow{2}{*}{{\mbox{$\mathop{\mathtt{\text{-}P(10)_{Llama}}}\limits$}\xspace}} 
& seen 
& 79.40 & 0.57 & 2.67 &  & 79.00 & 0.56 & 0.41 &  & 72.60 & 0.57 & 2.27 &  & 42.60 & 0.55 & 5.89 &  & 41.80 & 0.57 & \textbf{3.32} 
\\
 & unseen
  & \textbf{95.60} & 0.55 & 2.63 &  & \textbf{92.60} & 0.55 & 0.42 &  & \textbf{84.80} & 0.57 & 2.21 &  & \textbf{52.80} & 0.55 & 5.67 &  & \textbf{51.60} & 0.55 & 2.96 
\\
\bottomrule
\end{tabular}

\begin{tablenotes}[normal,flushleft]
\footnotesize
\item 
`Seen' and `unseen' indicate whether models are evaluated using instructions included during training or entirely novel instructions, respectively.
$^\uparrow$ and $^\downarrow$ indicate whether higher or lower values of the corresponding metric are preferable.
Within each row block, the best-performing model is highlighted in bold if the performance difference exceeds 5\%.
\par
\end{tablenotes}

\end{threeparttable}
\end{small}
\vspace{-10pt}
\end{table*}





\end{document}